\documentclass{article}
\PassOptionsToPackage{numbers, compress}{natbib}
\usepackage[final,main]{neurips_2026}
\usepackage[utf8]{inputenc}
\usepackage[T1]{fontenc}
\usepackage{hyperref}
\usepackage{url}
\usepackage{graphicx}
\usepackage{wrapfig}
\usepackage{float}
\usepackage{placeins}
\usepackage{booktabs}
\usepackage{array}
\usepackage{amsfonts}
\usepackage{nicefrac}
\usepackage{microtype}
\usepackage{xcolor}
\usepackage{pifont}
\usepackage{circledsteps}
\usepackage{algorithm}
\usepackage{algpseudocode}
\usepackage{capt-of}

\newcolumntype{C}[1]{>{\centering\arraybackslash}m{#1}}
\makeatletter
\newenvironment{breakablealgorithm}
  {%
   \par\refstepcounter{algorithm}%
   \hrule height 0.8pt depth 0pt \kern 2pt%
   \renewcommand{\caption}[2][\relax]{%
     {\raggedright\textbf{\fname@algorithm~\thealgorithm.} ##2\par}%
     \ifx\relax##1\relax
       \addcontentsline{loa}{algorithm}{\protect\numberline{\thealgorithm}##2}%
     \else
       \addcontentsline{loa}{algorithm}{\protect\numberline{\thealgorithm}##1}%
     \fi
     \kern 2pt\hrule\kern 2pt%
   }%
  }
  {\kern 2pt\hrule\par}
\makeatother

\title{DiversePlace: Diversity-Seeking Curriculum Reinforcement Learning for Macro Placement}

\author{%
  Wenrui Zhou\thanks{Equal contribution.}\quad \ \
  Jiashun Liu\textsuperscript{\ensuremath{*}}\quad
  Wenji Fang\quad
  Zhiyao Xie\quad
  Ling Pan\thanks{Corresponding author.}\\ [1.2ex]
  HKUST
}

\hypersetup{
  pdftitle={DiversePlace: Diversity-Seeking Curriculum Reinforcement Learning for Macro Placement},
  pdfauthor={Wenrui Zhou, Jiashun Liu, Wenji Fang, Zhiyao Xie, Ling Pan}
}

\begin{document}
\raggedbottom

\maketitle

\begin{abstract}
Macro placement is a critical early-stage decision in physical design because macro locations strongly constrain subsequent implementation stages.
The task is difficult because these high-impact discrete decisions must be made before downstream quality can be measured accurately, forcing training to rely on fast but imperfect training-time oracles such as half-perimeter wirelength (HPWL).
Existing RL-based placers therefore mostly optimize single HPWL-driven solutions, and their long sequential horizons often require heavy search assistance or offline expert data while leaving little room to preserve multiple competitive layouts.
We address this gap by making quality-controlled diversity an explicit objective for macro placement.
Instead of relying on a single proxy-optimal solution, we maintain a compact archive of HPWL-competitive yet meaningfully different placements for downstream selection.
We realize this idea in DiversePlace, a curriculum proximal policy optimization (PPO) framework that combines a quality-gated diversity reward with an expanding placement horizon, enabling from-scratch RL optimization while keeping exploration inside competitive HPWL regions.
On the ISPD2005 benchmarks, DiversePlace consistently improves top-5, top-10, and top-20 archive diversity and achieves lower or virtually identical HPWL on all eight designs with better efficiency.
\mbox{Code: \url{https://github.com/William030422/NIPS2026-DiversePlace}.}
\end{abstract}

\section{Introduction}
Macro placement is a decisive early step in modern chip design.
The macro locations strongly constrain subsequent standard-cell placement, routing, congestion optimization, and timing closure~\citep{markov2015progress,mirhoseini2021graph}. 
The critical challenge of macro placement stems from a large legality-constrained combinatorial search space with strong path dependence, where each macro decision reshapes the feasible space for all later decisions.
A line of research approached macro placement through optimization and supervised learning.
However, gradient-free black-box search~\citep{shi2023wiremask} is costly on large circuits, analytical relaxation~\citep{cheng2019replace, lin2021dreamplace} weakens alignment with discrete macro legality and tool-dependent objectives, and supervised learning-driven placers rely on expensive expert layouts~\citep{mirhoseini2021graph,lai2023chipformer,lee2025diffusion}.
Reinforcement learning (RL) naturally fits macro placement by training a policy to build layouts sequentially on a canvas while maximizing expected long-term rewards, thereby amortizing search over training without relying on expert datasets and achieving promising progress in recent placers.

\begin{figure}[t]
    \centering
    \includegraphics[width=\textwidth]{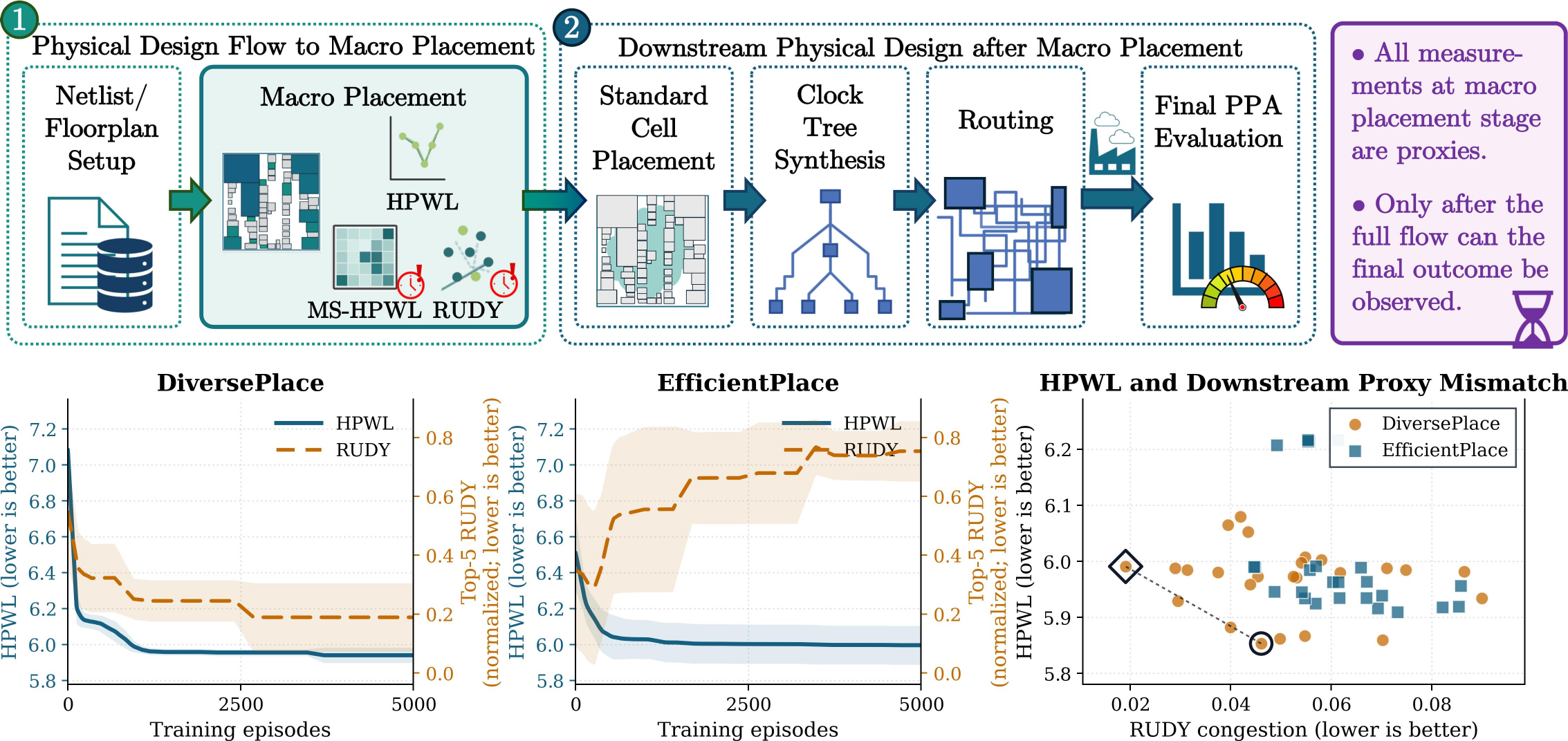}

\caption{
\textbf{Proxy-driven macro placement benefits from diverse candidate archives.}
\textbf{Top:} Macro placement is optimized before downstream outcomes such as routing and final PPA can be accurately measured, making placement-stage objectives imperfect proxies.
\textbf{Bottom left/middle:} On \textit{adaptec1}, we evaluate five independent runs (seeds 0--4). Each seed's HPWL-ranked top-5 archive undergoes mixed-size placement and RUDY congestion evaluation. DiversePlace, trained without downstream feedback, yields archives whose best RUDY improves as HPWL decreases, while the similar-mode EfficientPlace archive lacks consistent proxy-to-congestion transfer.
\textbf{Bottom right:} Best-HPWL and best-RUDY placements do not always coincide. We show that diverse HPWL-competitive archives improve the chance of selecting layouts strong under both HPWL and downstream congestion proxies.
}
    \label{fig:intro-proxy-gap}
\end{figure}

However, RL-based macro placement is inherently proxy-driven. 
The final quality of a macro placement is determined only after downstream implementation stages as illustrated in Figure~\ref{fig:intro-proxy-gap}, which is unavailable during RL training in this stage. 
Therefore, existing RL-based placers typically optimize half-perimeter wirelength (HPWL), which is a proxy computed from pin bounding boxes before routed or signoff-oriented metrics are available~\citep{lin2021dreamplace,mirhoseini2021graph,lai2022maskplace,lai2023chipformer,geng2024efficientplace} (detailed in Appendix~\ref{app:hpwl-update-illustration}).
Despite its utility, it is an imperfect training-time oracle for downstream routability, timing, and power, performance, and area (PPA) 
~\citep{spindler2007rudy,cheng2021deeppr,kahng2021advancing,agnesina2023autodmp}.
The return-optimization nature of RL can therefore drive the policy to overoptimize such an imperfect proxy~\citep{gao2023scaling}.
Figure~\ref{fig:intro-proxy-gap} illustrates this mismatch on \textit{adaptec1}: HPWL improvement does not necessarily translate into improved downstream congestion measured by
rectangular uniform wire density (RUDY, detailed in Appendix~\ref{app:mixed-size-congestion})
under a recent RL-based state-of-the-art (SOTA) algorithm~\citep{geng2024efficientplace}.
Moreover, sole return optimization in standard RL with imperfect reward can concentrate the policy around one proxy-preferred layout mode~\citep{eysenbach2019diversity}, while discarding alternative placements that are suboptimal under HPWL but more favorable downstream.

This proxy mismatch exposes an overlooked challenge in RL-based macro placement: the loss of high-quality candidate diversity. 
Most existing RL-based placers focus on optimizing single-solution proxy quality, implicitly assuming that the best placement under the training reward is also the most useful downstream candidate~\citep{lai2022maskplace,lai2023chipformer, geng2024efficientplace}. 
However, under an imperfect oracle, this assumption can be fragile. 
If this mode later underperforms in downstream evaluation, nearby candidates may fail in similar ways, whereas distinct high-quality modes can expose different behaviors across the later physical-design flow~\cite{jain2023biologicalsequencedesigngflownets}.
\emph{We therefore argue that an RL placer should return not only one HPWL-minimizing placement, but also a compact archive of high-quality and meaningfully diverse macro placements for downstream selection.}

Motivated by this proxy mismatch, we pursue a diversity-seeking RL formulation for macro placement, where the agent discovers a compact set of HPWL-competitive yet meaningfully different layouts for downstream selection.
Realizing this formulation raises two coupled design challenges.
\noindent\textbf{Challenge I. Quality-controlled diversity under an imperfect oracle.}
Diversity should be quality-controlled as unconstrained novelty can produce physically irrelevant layouts, whereas overly tight HPWL-based selection can collapse the search to a few proxy-favorable modes and discard other downstream-viable candidates~\citep{markov2015progress,cheng2019replace,lin2021dreamplace,pugh2016qualitydiversity}.
Most existing learning-based placers overlook this tension by optimizing single-solution proxy quality rather than maintaining a diverse set of HPWL-competitive candidates under an imperfect oracle~\citep{mirhoseini2021graph,cheng2021deeppr,lai2022maskplace,lai2023chipformer,geng2024efficientplace}.
\noindent\textbf{Challenge II. Long-horizon sequential optimization.}
Macro placement is a long-horizon decision process in which each trajectory places many macros.
Early large-macro decisions reshape the canvas, constrain later feasible regions, and weaken credit assignment across the sequence~\citep{lai2022maskplace,geng2024efficientplace}.
This makes diversity-seeking optimization harder, since the agent must learn to produce competitive placements while exploring distinct high-quality layout modes across a large sequential search space.
Thus, the central challenge is not merely to add a diversity term, but to discover distinct layout modes inside a competitive HPWL region while keeping the RL optimization problem learnable.

To realize this formulation, we propose DiversePlace, a novel diversity-seeking curriculum proximal policy optimization (PPO) framework for macro placement~\citep{schulman2017ppo}.
To address quality-controlled diversity, DiversePlace grants diversity rewards only to HPWL-competitive rollouts, and stores their area-weighted macro-displacement descriptors in a quality-diversity (QD) map to encourage new high-quality layout modes~\citep{mouret2015illuminating,pugh2016qualitydiversity}.
To address long-horizon optimization, DiversePlace trains PPO with an expanding placement curriculum that gradually increases the number of macros controlled by the policy rather than exposing the full horizon from the start~\citep{bengio2009curriculum}.
These designs keep sequential optimization learnable while guiding PPO toward multiple HPWL-competitive placement modes.


We evaluate DiversePlace on ISPD2005 macro-placement benchmarks, following prior work~\cite{geng2024efficientplace,lai2022maskplace}.
Across eight designs, DiversePlace consistently improves top-5, top-10, and top-20 archive diversity over recent SOTA methods, while achieving lower or virtually identical HPWL at multiple checkpoints with better efficiency.
We further evaluate mixed-size placement and congestion, showing that the diverse archive increases the opportunity to select downstream-favorable candidates from a HPWL-ranked top-$k$ set.
Ablations validate both the quality-gated diversity objective and the expanding curriculum schedule.
The main contributions are summarized as follows.

\noindent\textbf{(1) Quality-controlled diversity for macro placement.}
We address critical, overlooked diversity loss in RL-based macro placement with a novel QD learning mechanism that encourages multiple diverse, high-quality layouts as competitive candidates for downstream physical-design tools.

\noindent\textbf{(2) Curriculum training for diversity-seeking RL for efficient placement optimization from scratch.}
We introduce a curriculum PPO framework that progressively expands the placement horizon, enabling efficient short-to-long optimization with faster early HPWL improvement and no loss in final placement quality.

\noindent\textbf{(3) Comprehensive validation and analysis.}
We extensively analyze optimization strategies, learning designs, and placement diversity, offering practical insights for future method development.

\section{Related Work}

\paragraph{Optimization and Generative Placement Methods.}
Classical placement methods span partitioning, stochastic floorplanning, black-box search, and analytical optimization, exposing a long-standing trade-off among feasibility, objective flexibility, and scalability~\citep{markov2015progress}. 
Stochastic floorplanners with sequence-pair or B$^\ast$-tree representations and black-box methods such as WireMask-EA directly optimize discrete macro or mixed-size layouts, but their perturb-and-evaluate loops scale poorly on large circuits~\citep{kirkpatrick1983optimization,murata1996sequencepair,chang2000btree,shi2023wiremask}. 
Analytical placers such as RePlAce and DREAMPlace scale through continuous optimization of smoothed wirelength and density objectives, but weaken alignment with discrete macro actions, hard legality masks, and non-differentiable tool-dependent rewards~\citep{cheng2019replace,lin2021dreamplace}. 
Generative placers, including Flora, GraphPlanner, GAN-Place, and diffusion-based chip placement, learn layout priors from existing or synthetic designs, but depend on data coverage, supervision fidelity, and generalization to unseen netlists, technologies, and constraints~\citep{liu2022floorplanning,liu2023graphplanner,lu2024ganplace,lee2025diffusion}. 
These limitations motivate our focus on RL-based macro placement.

\paragraph{RL-Based Methods.}
RL-based methods are prominent because macro placement naturally admits a sequential MDP formulation \citep{mirhoseini2021graph,cheng2021deeppr}. AlphaChip introduced this view, and DeepPR and PRNet extended it toward placement-routing pipelines, but these early methods remained slow and often weak in legality or scalability \citep{mirhoseini2021graph,cheng2021deeppr,cheng2022prnet,lai2022maskplace}. MaskPlace improved efficiency with pixel-level masks and dense rewards, and ChiPFormer reduced online cost through offline pretraining, yet both still bear the burden of long-horizon sequential modeling, with the latter additionally depending on offline expert coverage \citep{lai2022maskplace,lai2023chipformer}. EfficientPlace further strengthened the quality-runtime trade-off with tree-search-guided PPO, but much of the gain comes from MCTS and sequential freezing rather than from the RL policy alone \citep{geng2024efficientplace}. More broadly, this line is still largely driven by HPWL or macro-HPWL proxies, while timing and routability are typically evaluated only after placement \citep{mirhoseini2021graph,lai2022maskplace,lai2023chipformer,geng2024efficientplace,shi2023wiremask}. 
This calls for a cleaner RL-native route that better exploits the strengths of policy learning without tying the agent too tightly to a single proxy reward, thereby leaving more room for stronger overall placement quality across later flow stages.

\section{Preliminaries}

\paragraph{Macro placement and HPWL.}
Macro placement assigns a set of hard macros $\mathcal{M}=\{m_1,\ldots,m_T\}$ to legal locations on a chip canvas under boundary and non-overlap constraints. These early macro locations strongly constrain later standard-cell placement, routing, congestion optimization, and timing closure. Since routed wirelength and signoff-oriented objectives are too expensive to evaluate during RL training, learning-based placers commonly optimize half-perimeter wirelength (HPWL) as a fast placement-stage proxy. For a net $e$ with pin set $\mathcal{P}_e$, HPWL is the half-perimeter of the bounding box covering its pins:
\[
\mathrm{HPWL}
=
\sum_{e\in\mathcal{E}}
\left(
\max_{p\in\mathcal{P}_e} p_x-\min_{p\in\mathcal{P}_e} p_x
+
\max_{p\in\mathcal{P}_e} p_y-\min_{p\in\mathcal{P}_e} p_y
\right).
\]
Appendix~\ref{app:hpwl-update-illustration} illustrates how adding one macro updates the affected HPWL bounding boxes.

\paragraph{Macro placement as sequential decision making.}
Macro placement can be viewed as a finite-horizon sequential decision-making problem in which a layout is constructed one macro at a time. 
At step $t$, the environment contains a partial placement of macros $m_1,\ldots,m_{t-1}$. 
The state $s_t$ encodes this partial layout together with canvas occupancy, connectivity information, and legality constraints~\cite{lai2022maskplace}. 
The action $a_t=(x_t,y_t)$ selects a legal grid location for the current macro $m_t$; invalid locations are masked out by boundary and non-overlap constraints. 
After the action is taken, the environment deterministically fixes $m_t$ at the selected location, updates the canvas and legality masks, and proceeds to the next macro. 
After $T$ decisions, the trajectory yields a complete macro placement. 
The reward is derived from placement quality, typically through HPWL or its incremental change during placement. 
We use PPO as the policy optimizer.
\section{Method}

\subsection{Motivation and Overview}
\begin{wrapfigure}{r}{0.54\textwidth}
    \centering
    \includegraphics[width=\linewidth]{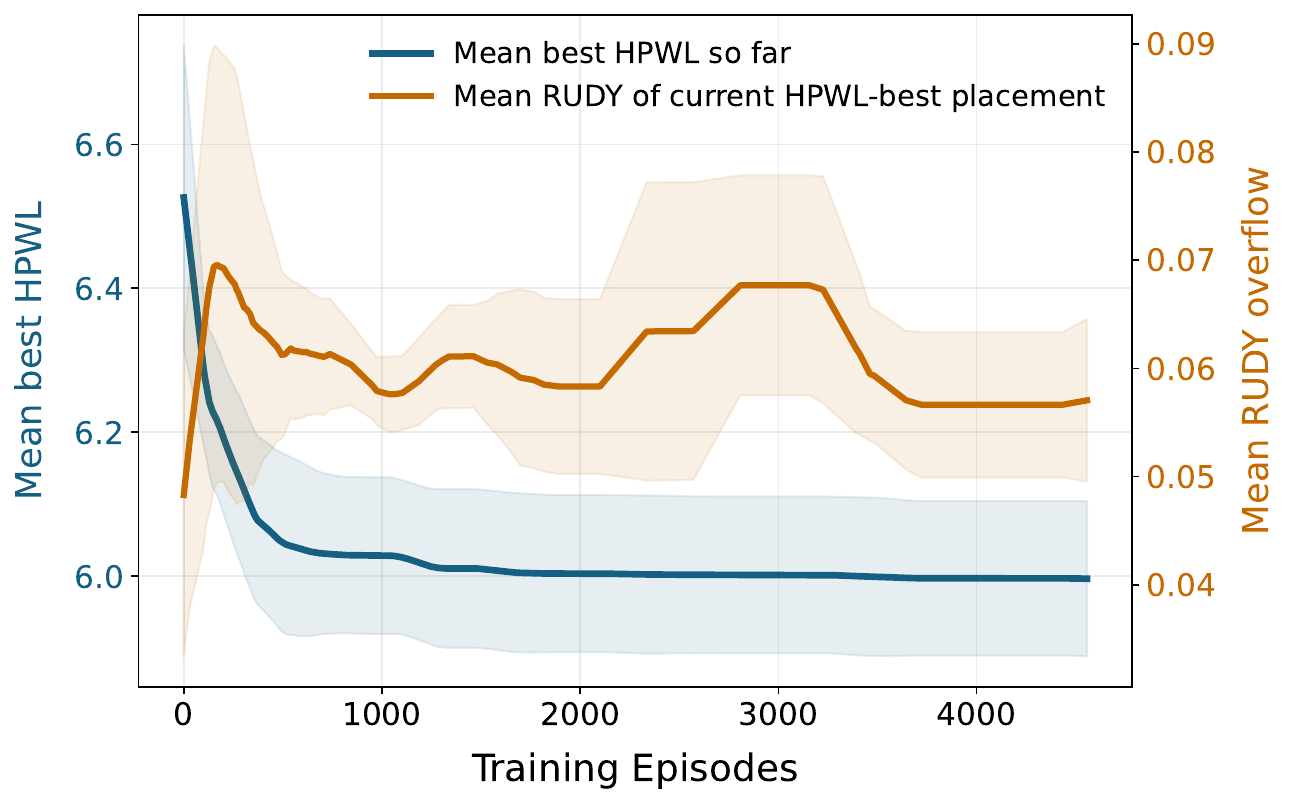}
    \caption{\textbf{Best HPWL need not imply the best downstream proxy.}
    Across five seeds (0--4) on Adaptec1, the current SOTA algorithm has non-monotonic mean RUDY overflow at its current HPWL-best placement.}
    \label{fig:method-proxy-gap}
\end{wrapfigure}

DiversePlace is motivated by a training-time oracle mismatch where HPWL serves as a useful training signal but is imperfect as a design target.
HPWL is cheap to evaluate online and often correlates with physical-design quality, but downstream stages such as routing, congestion optimization, and timing closure can favor placements that are not the single HPWL winner.
Therefore, the HPWL-best placement does not necessarily yield the best downstream outcome, as shown in Figure~\ref{fig:method-proxy-gap}. 
This motivates a placer that returns multiple HPWL-competitive layouts rather than a single proxy-optimal solution from an imperfect oracle.
However, diversity is meaningful only when it remains coupled with quality.
A reward that favors difference alone can be satisfied by poor or physically irrelevant placements, whereas HPWL-only optimization tends to collapse to a narrow set of proxy-favorable modes.
The goal is therefore to discover multiple distinct layout basins in a competitive HPWL region rather than trade HPWL for superficial spread.

DiversePlace addresses this problem with a quality-gated diversity objective and a curriculum that preserves optimization quality over long horizons. 
First, in each curriculum window, PPO runs a warmup phase under the base placement reward to establish a local HPWL baseline and collect high-quality reference placements.
Second, the diversity phase grants a capped terminal bonus only to rollouts whose final HPWL remains within a small tolerance of that baseline, preventing novelty search from drifting into low-quality regions.
For each quality-eligible rollout, the bonus is computed from area-weighted, size-normalized macro displacement relative to the nearest high-quality reference and inserted into a two-dimensional quality-diversity (QD) map, whose cells retain the best-HPWL elite for each movement pattern.
Third, to keep this objective stable as the horizon grows, the next window expands the PPO-controlled horizon, reuses only a short prefix bridge selected from top placements in the previous window, and completes the remaining suffix greedily so every rollout is still evaluated as a full placement.
This design lets PPO explore alternative layout basins while remaining anchored to placements that are still competitive under the imperfect reward signal.

\subsection{Diversity-Seeking Reward}
\label{sec:method-diversity-reward}
\paragraph{Quality-gated references.}
DiversePlace applies diversity to completed placements rather than partial rollouts. In each episode, PPO places the macros currently exposed by the curriculum, the remaining suffix is completed greedily, and the resulting full placement is evaluated once at the terminal transition. Diversity is meaningful only among competitive placements. Therefore, at each curriculum window $k$, DiversePlace first runs a warmup phase under the base placement reward and records the best warmup placement $P_k^\star$. It then forms a reference set $\mathcal{R}_k$ from warmup placements whose HPWL lies within a small tolerance of this best value:
\begin{equation}
\mathcal{R}_k =
\left\{
P : \mathrm{HPWL}(P) \le (1+\epsilon_q)\mathrm{HPWL}(P_k^\star)
\right\}.
\end{equation}
During the subsequent diversity phase, a rollout outside this HPWL band receives no diversity bonus, even if it differs substantially from previous placements. This gate restricts novelty search to a local region of placements that remain competitive in proxy quality.

\begin{wrapfigure}{r}{0.48\textwidth}
    \centering
    \includegraphics[width=\linewidth]{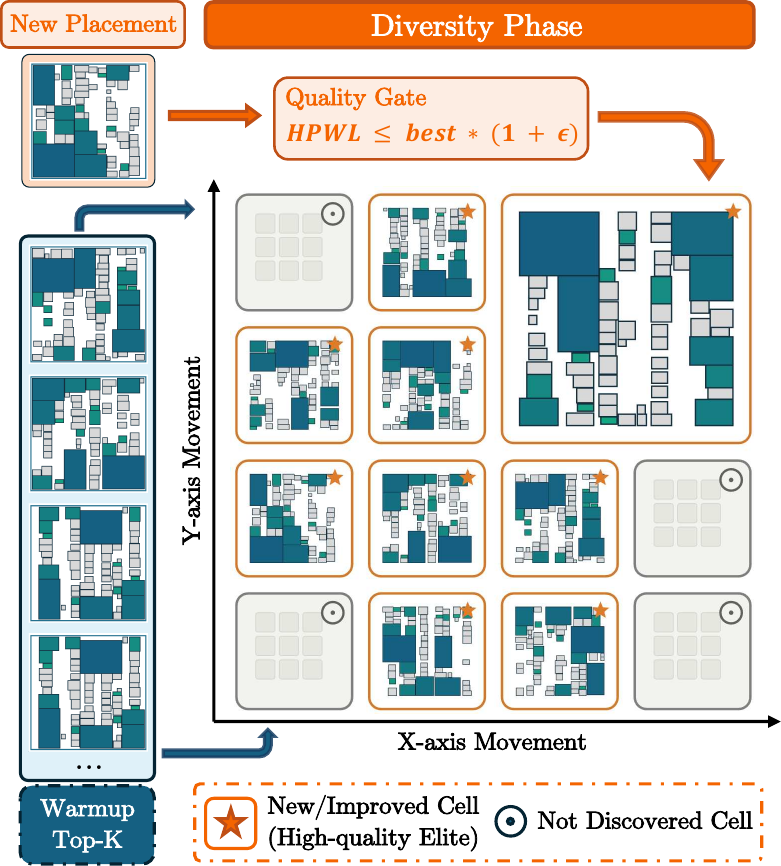}
    \caption{\textbf{Quality-gated diversity.}
    HPWL-gated candidate placements enter a QD map parameterized by area-weighted, size-normalized displacement from the nearest high-quality reference.}
    \label{fig:method-diversity}
\end{wrapfigure}
\paragraph{Normalized displacement.}
Given a completed placement $P$ and a reference placement $R\in\mathcal{R}_k$, DiversePlace measures novelty from the movement of macros in the current diversity scope $\mathcal{S}_k$. For macro $m_i$ with placement coordinates $(x_i,y_i)$ in $P$, reference coordinates $(x_i^R,y_i^R)$ in $R$, and size $(w_i,h_i)$, the normalized displacement is
\begin{equation}
d_i(P,R)=
\frac{
\sqrt{(x_i-x_i^R)^2+(y_i-y_i^R)^2}
}{
\sqrt{w_i^2+h_i^2}
}.
\end{equation}
We aggregate these displacements with area weights $\omega_i=w_i h_i/\sum_j w_jh_j$:
\begin{equation}
D(P,R)=\sum_{i\in\mathcal{S}_k}\omega_i d_i(P,R).
\end{equation}
Diagonal normalization makes movement comparable across macro sizes, while area weighting gives larger macros more influence because they shape the global placement basin more strongly.

\paragraph{QD descriptor and elite map.}
DiversePlace compares each placement with its nearest high-quality reference,
\begin{equation}
R^\star(P)=\arg\min_{R\in\mathcal{R}_k} D(P,R),
\end{equation}
so novelty is measured relative to the most similar competitive mode rather than to an arbitrary reference. From $R^\star(P)$, DiversePlace builds a two-dimensional descriptor that separates movement along the two canvas axes:
\begin{equation}
b(P)=
\left(
g\!\left(\sum_{i\in\mathcal{S}_k}\omega_i
\frac{|x_i-x_i^{R^\star(P)}|}{\sqrt{w_i^2+h_i^2}}\right),
g\!\left(\sum_{i\in\mathcal{S}_k}\omega_i
\frac{|y_i-y_i^{R^\star(P)}|}{\sqrt{w_i^2+h_i^2}}\right)
\right),
\end{equation}
where $g(z)=z/(1+z)$ maps each coordinate to a bounded range. The descriptor is inserted into a two-dimensional QD map whose bin edges are fit within each window from the warmup descriptor distribution. Each cell stores the lowest-HPWL elite discovered for that movement pattern. Empty cells represent unexplored layout modes, whereas occupied cells retain the best placement found in that region of the descriptor space.

\paragraph{Terminal diversity bonus.}
The QD map changes training through the terminal PPO reward. If placement $P$ passes the quality gate, its raw diversity bonus is
\begin{equation}
r_{\mathrm{div}}(P)=
\lambda D(P,R^\star(P))\,\eta(c(P)),
\end{equation}
where $c(P)$ is the QD cell of $P$ and
\begin{equation}
\eta(c)=
\left\{
\begin{array}{ll}
1, & \mbox{if } c \mbox{ is empty or } P \mbox{ improves the cell elite},\\
(1+n_c)^{-1/2}, & \mbox{otherwise}.
\end{array}
\right.
\end{equation}
with $n_c$ the visit count of cell $c$. The final bonus is capped relative to the terminal placement reward:
\begin{equation}
\tilde r_{\mathrm{div}}(P)=
\min\left(r_{\mathrm{div}}(P),\,\beta |r_{\mathrm{term}}|\right).
\end{equation}
The terminal PPO transition stores $r_{\mathrm{term}}+\tilde r_{\mathrm{div}}$, while intermediate rewards remain unchanged. Therefore, the QD map is not merely an evaluation archive. It directly biases PPO toward quality-eligible placements that enter new cells or improve existing elites.

\subsection{Curriculum over Placement Horizons}
\label{subsec:curriculum}
\begin{figure}[H]
    \centering
    \includegraphics[width=\linewidth]{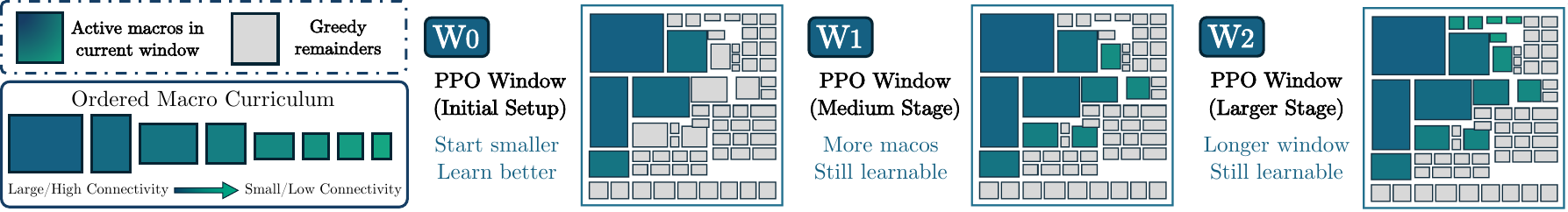}
    \caption{\textbf{Curriculum over placement horizons.}
    DiversePlace starts from a short PPO-controlled macro window and progressively expands it, while greedily completing the remaining macros so every rollout is evaluated as a full placement.}
    \label{fig:method-curriculum}
\end{figure}

\paragraph{Why curriculum is needed.}
The diversity reward is useful only if PPO can first reach a competitive HPWL region.
This is difficult when the policy must solve the full macro-placement horizon from the beginning, because early macro decisions determine the feasible basin for many later placements and make credit assignment increasingly weak. 
DiversePlace therefore uses curriculum learning to make quality-gated diversity compatible with long-horizon RL optimization.

\paragraph{Expanding horizon.}
Let $T$ be the total number of macros and let $H_k$ be the number of macros controlled by PPO in curriculum window $k$. DiversePlace starts from a small horizon $H_0$ and expands it by a fixed increment:
\begin{equation}
H_k=\min(T, H_0+k\Delta H).
\end{equation}
Within window $k$, PPO places macros $m_1,\dots,m_{H_k}$ using the visual state, legal-position mask, and wire mask. The remaining macros $m_{H_k+1},\dots,m_T$ are then placed greedily so that every episode still produces a complete placement and a full HPWL value. After each window, the horizon expands and the same actor-critic networks continue training. This schedule reduces the initial credit-assignment burden while preserving the full-placement objective at every episode.

\paragraph{Window schedule.}
Each curriculum window has two phases. The warmup phase optimizes the base placement reward and establishes a local HPWL baseline together with the reference set for the current horizon. The diversity phase then searches for alternative placements within that local quality region using the QD bonus defined above. Because the gate and the references are recomputed at every horizon, the diversity signal stays aligned with the difficulty of the current window instead of relying on stale placements from earlier, shorter horizons.

\paragraph{Prefix bridge.}
At horizon boundaries, DiversePlace ranks completed placements from the previous window by HPWL and retains a small top-$K$ prefix pool. 
The next window reuses only the first few macro locations from one of these placements as a short prefix bridge. 
This transfer stabilizes early high-impact decisions that largely determine the feasible basin for later macros, while leaving most of the newly exposed horizon free for PPO refinement. 
The transferred segment is deliberately small (e.g., two macros), allowing the policy to still reshape the placement as the horizon expands.
Together, the expanding horizon, per-window warmup-then-diversity schedule, and short prefix bridge keep PPO in a competitive HPWL region with enough freedom to discover multiple layout modes.

\section{Experiments}
\label{sec:experiments}

\subsection{Benchmarks and Settings}
We evaluate DiversePlace against one classical reference baseline, simulated annealing (SA)~\citep{kirkpatrick1983optimization}, and five RL-based placers including GraphPlace~\citep{mirhoseini2021graph}, DeepPR~\citep{cheng2021deeppr}, MaskPlace~\citep{lai2022maskplace}, ChiPFormer~\citep{lai2023chipformer}, and EfficientPlace~\citep{geng2024efficientplace}.
All runs in this work, including the EfficientPlace reruns and all DiversePlace experiments, are conducted on a single server with two Intel Xeon Platinum 8375C CPUs and eight NVIDIA RTX 3090 GPUs, each with 24 GB of memory; each seed uses one GPU.
Consistent with recent macro-placement studies~\citep{lai2022maskplace, lai2023chipformer, geng2024efficientplace}, we evaluate all methods on the widely used ISPD2005 benchmark suite~\citep{nam2005ispd}, which contains the eight instances \textit{adaptec1--4} and \textit{bigblue1--4}.

\subsection{Evaluation Metrics}
We use HPWL, reported in units of $10^5$, as the primary placement-quality metric; lower values are better.
For comparisons with the matching budget, we report the cumulative-best HPWL after global episodes of 1K, 2K, and 10K.
To assess whether a method preserves multiple competitive placement modes, we also analyze the global top-$k$ solution archive saved during training.
For archive diversity, we reuse the area-weighted, size-normalized macro-displacement defined in Section~\ref{sec:method-diversity-reward}, but apply it pairwise between complete placements over their shared comparable macro set.
For an archive $\{P_i\}_{i=1}^k$, the diversity score is the mean pairwise distance
\begin{equation}
D_k=\frac{2}{k(k-1)}\sum_{i<j}d(P_i,P_j).
\end{equation}
This pairwise-archive view follows standard set-diversity practice in quality-diversity methods~\cite{mouret2015illuminating,pugh2016qualitydiversity}.
Thus, $D_k$ is a unitless measure of geometric variation among high-quality placements; higher values indicate a broader set of distinct layouts.

\subsection{Results and Analysis}

\begin{table*}[!t]
\centering
\scriptsize
\setlength{\tabcolsep}{0.5pt}
\caption{
HPWL values ($\times 10^5$) obtained by compared macro-placement methods on eight ISPD2005 circuits~\cite{nam2005ispd}. 
Results for prior methods are taken from the corresponding reported sources~\cite{mirhoseini2021graph,cheng2021deeppr,lai2022maskplace,lai2023chipformer,geng2024efficientplace}, whereas EfficientPlace and DiversePlace are evaluated using a single predetermined batch of five independent runs with seeds 0--4, all of which are included when reporting means and sample standard deviations.
Numbers after method names give the global-episode counts.
For the 1K and 2K rows, runtime in hours is reported below HPWL in parentheses. Lower HPWL and runtime are better; \textbf{bold} and \underline{underlining} mark the best and second-best HPWL results.
}
\label{tab:hpwl-overall}
\resizebox{\textwidth}{!}{%
\begin{tabular}{@{}C{2cm}*{8}{C{1.5cm}}@{}}
\toprule
Method & adaptec1 & adaptec2 & adaptec3 & adaptec4 & bigblue1 & bigblue2 & bigblue3 & bigblue4 \\
\midrule
SA (1M) & 19.24$\pm$1.42 & 102.84$\pm$5.83 & 140.34$\pm$7.48 & 148.57$\pm$23.08 & 5.92$\pm$0.98 & 12.09$\pm$0.42 & 270.46$\pm$43.76 & 248.38$\pm$12.25 \\
GraphPlace (50K) & 30.01$\pm$2.98 & 351.71$\pm$38.20 & 358.18$\pm$13.95 & 151.42$\pm$9.72 & 10.58$\pm$1.29 & 14.78$\pm$0.95 & 357.48$\pm$47.83 & 440.70$\pm$15.95 \\
DeepPR (3K) & 19.91$\pm$2.13 & 203.51$\pm$6.27 & 347.16$\pm$4.32 & 311.86$\pm$56.74 & 23.33$\pm$3.65 & 11.38$\pm$0.20 & 430.48$\pm$12.18 & 433.90$\pm$5.26 \\
MaskPlace (3K) & 7.62$\pm$0.67 & 75.16$\pm$4.97 & 100.24$\pm$13.54 & 87.99$\pm$3.25 & 3.04$\pm$0.06 & 5.75$\pm$0.11 & 90.04$\pm$4.83 & 103.26$\pm$2.69 \\
ChiPFormer (2K) & 6.62$\pm$0.05 & 67.10$\pm$5.46 & 76.70$\pm$1.15 & 68.80$\pm$1.59 & 2.95$\pm$0.04 & 5.44$\pm$0.10 & 72.92$\pm$2.56 & 102.84$\pm$0.15 \\
\midrule
EfficientPlace (1K) & \shortstack{6.03$\pm$0.12\\(0.82h)} & \shortstack{47.19$\pm$3.50\\(0.87h)} & \shortstack{57.42$\pm$1.53\\(1.72h)} & \shortstack{61.32$\pm$1.30\\(4.25h)} & \shortstack{2.29$\pm$0.03\\(0.91h)} & \shortstack{4.75$\pm$0.13\\(0.37h)} & \shortstack{74.96$\pm$21.93\\(3.20h)} & \shortstack{89.88$\pm$1.95\\(2.29h)} \\
DiversePlace (1K) & \shortstack{6.01$\pm$0.09\\(0.45h)} & \shortstack{46.81$\pm$0.68\\(0.63h)} & \shortstack{56.93$\pm$0.74\\(1.24h)} & \shortstack{59.64$\pm$0.84\\(2.88h)} & \shortstack{2.31$\pm$0.02\\(0.95h)} & \shortstack{4.78$\pm$0.12\\(0.19h)} & \shortstack{73.10$\pm$11.92\\(2.61h)} & \shortstack{88.60$\pm$2.28\\(1.79h)} \\
EfficientPlace (2K) & \shortstack{6.00$\pm$0.12\\(1.47h)} & \shortstack{46.40$\pm$3.54\\(1.65h)} & \shortstack{57.29$\pm$1.68\\(3.33h)} & \shortstack{60.57$\pm$1.44\\(8.50h)} & \shortstack{\underline{2.29$\pm$0.03}\\(1.70h)} & \shortstack{4.69$\pm$0.13\\(0.69h)} & \shortstack{71.43$\pm$23.44\\(6.23h)} & \shortstack{88.33$\pm$1.78\\(4.51h)} \\
DiversePlace (2K) & \shortstack{\underline{5.99$\pm$0.07}\\(0.91h)} & \shortstack{\underline{44.59$\pm$1.15}\\(1.28h)} & \shortstack{\underline{55.48$\pm$0.54}\\(2.51h)} & \shortstack{\underline{58.66$\pm$0.89}\\(5.64h)} & \shortstack{2.29$\pm$0.02\\(2.45h)} & \shortstack{4.70$\pm$0.04\\(0.41h)} & \shortstack{\underline{61.81$\pm$15.64}\\(5.19h)} & \shortstack{\underline{85.40$\pm$2.53}\\(3.50h)} \\
EfficientPlace (10K) & 6.00$\pm$0.12 & 46.12$\pm$3.45 & 56.46$\pm$1.52 & 59.19$\pm$1.46 & \textbf{2.28$\pm$0.02} & \underline{4.57$\pm$0.08} & 68.65$\pm$22.07 & 85.71$\pm$1.49 \\
DiversePlace (10K) & \textbf{5.91$\pm$0.05} & \textbf{42.44$\pm$0.87} & \textbf{55.25$\pm$0.66} & \textbf{56.40$\pm$0.92} & 2.29$\pm$0.02 & \textbf{4.56$\pm$0.05} & \textbf{59.48$\pm$14.62} & \textbf{82.91$\pm$2.70} \\
\bottomrule
\end{tabular}%
}
\par\medskip
\centering
\scriptsize
\setlength{\tabcolsep}{0.5pt}
\caption{
Top-$K$ archive quality and diversity comparison. 
Each cell reports DiversePlace / EfficientPlace, with the winning value highlighted in \textbf{bold}. 
HPWL rows report archive quality, where lower is better. $D_k^{\mathrm{full}}$ reports mean pairwise normalized macro-displacement over complete placements. $D_k^{\mathrm{PPO}}$ removes the WireMask greedy tail and evaluates only the PPO-controlled macro range.}
\label{tab:diversity}
\resizebox{\textwidth}{!}{%
\begin{tabular}{@{}C{1.15cm}C{0.4cm}*{8}{C{1.5cm}}@{}}
\toprule
Metric & $k$ & adaptec1 & adaptec2 & adaptec3 & adaptec4 & bigblue1 & bigblue2 & bigblue3 & bigblue4 \\
\midrule
HPWL & 5 & \textbf{5.96} / 6.00 & \textbf{42.77} / 46.24 & \textbf{55.82} / 56.87 & \textbf{57.16} / 59.88 & 2.29 / 2.29 & 4.61 / 4.61 & \textbf{60.92} / 69.75 & \textbf{85.05} / 86.79 \\
HPWL & 10 & \textbf{5.98} / 6.00 & \textbf{42.92} / 46.33 & \textbf{56.01} / 57.05 & \textbf{57.58} / 60.21 & 2.29 / 2.29 & 4.63 / \textbf{4.62} & \textbf{61.77} / 70.32 & \textbf{86.03} / 87.28 \\
HPWL & 20 & \textbf{6.00} / 6.01 & \textbf{43.08} / 46.42 & \textbf{56.25} / 57.24 & \textbf{58.00} / 60.55 & 2.30 / \textbf{2.29} & 4.66 / \textbf{4.64} & \textbf{62.75} / 70.98 & \textbf{86.89} / 87.90 \\
\midrule
$D_k^{\mathrm{full}}$ & 5 & \textbf{3.98} / 2.13 & \textbf{1.83} / 1.48 & \textbf{4.20} / 3.41 & \textbf{10.89} / 8.56 & \textbf{3.80} / 2.79 & \textbf{2.28} / 0.62 & \textbf{4.57} / 3.37 & \textbf{6.13} / 4.42 \\
$D_k^{\mathrm{full}}$ & 10 & \textbf{3.89} / 2.17 & \textbf{1.77} / 1.51 & \textbf{4.28} / 3.62 & \textbf{11.42} / 8.75 & \textbf{3.82} / 2.76 & \textbf{2.20} / 0.62 & \textbf{4.66} / 3.38 & \textbf{6.24} / 4.27 \\
$D_k^{\mathrm{full}}$ & 20 & \textbf{3.86} / 2.17 & \textbf{1.75} / 1.53 & \textbf{4.34} / 3.68 & \textbf{11.69} / 8.82 & \textbf{3.84} / 2.71 & \textbf{2.49} / 0.67 & \textbf{4.65} / 3.47 & \textbf{6.26} / 4.28 \\
\midrule
$D_k^{\mathrm{PPO}}$ & 5 & \textbf{1.87} / 0.01 & \textbf{0.34} / 0.01 & \textbf{0.36} / 0.02 & \textbf{0.68} / 0.19 & \textbf{1.10} / 0.15 & \textbf{2.32} / 0.13 & \textbf{0.86} / 0.10 & \textbf{0.70} / 0.40 \\
$D_k^{\mathrm{PPO}}$ & 10 & \textbf{1.80} / 0.02 & \textbf{0.27} / 0.01 & \textbf{0.40} / 0.11 & \textbf{0.71} / 0.19 & \textbf{1.12} / 0.11 & \textbf{2.22} / 0.10 & \textbf{0.85} / 0.10 & \textbf{0.69} / 0.30 \\
$D_k^{\mathrm{PPO}}$ & 20 & \textbf{1.81} / 0.01 & \textbf{0.26} / 0.01 & \textbf{0.38} / 0.12 & \textbf{0.76} / 0.24 & \textbf{1.13} / 0.06 & \textbf{2.54} / 0.13 & \textbf{0.84} / 0.14 & \textbf{0.72} / 0.29 \\
\bottomrule
\end{tabular}%
}
\end{table*}

\paragraph{HPWL Results.}
Table~\ref{tab:hpwl-overall} presents the overall HPWL comparison against prior macro-placement methods.
Under a controlled protocol, DiversePlace is faster than EfficientPlace on seven of the eight benchmarks at 1K episodes.
It also delivers lower or essentially comparable HPWL on most datasets, achieving the best 10K HPWL on seven benchmarks.

The presented checkpoints in Table~\ref{tab:hpwl-overall} and full trajectories in Appendix Figure~\ref{fig:hpwl-trajectory} show that DiversePlace's advantage emerges early rather than only during late-stage training.
We argue that placement quality is largely determined by the early basin reached by the policy: PPO mainly steers search toward stronger initial basins, after which greedy placement induced by WireMask and DreamPlace refinement can recover much of the remaining quality~\cite{lai2022maskplace}.
By 1K and 2K episodes, DiversePlace already achieves lower HPWL than EfficientPlace on six of eight benchmarks, increasing this count to seven by 10K episodes.
Moreover, by incorporating curriculum, we further show that this early quality does not require a larger wall-clock budget on most benchmarks: DiversePlace uses $0.73\times$ the 1K runtime and $0.81\times$ the 2K runtime of EfficientPlace on average.
These results indicate that our paradigm improves early solution quality while preserving practical training efficiency.

\begin{figure*}[ht]
\centering
\includegraphics[width=\textwidth]{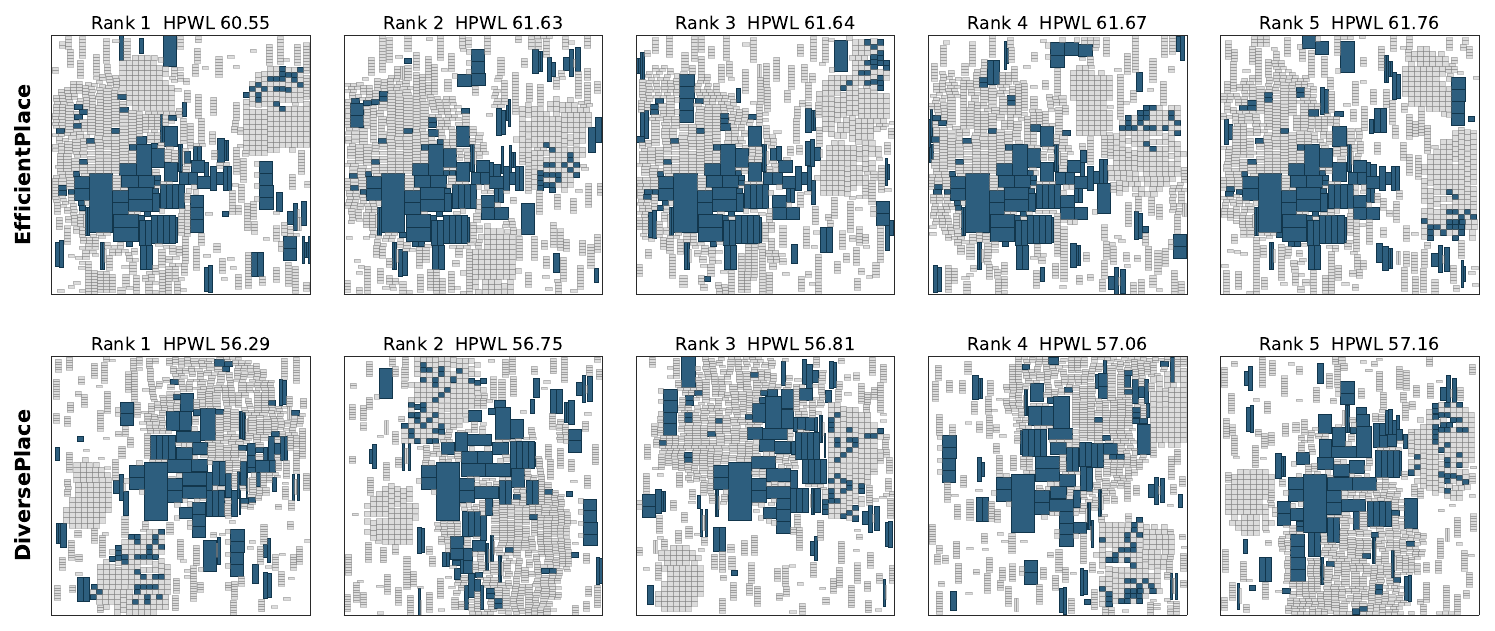}
\caption{
\textbf{Top-5 archive placements on \textit{adaptec4}.} We compare EfficientPlace and DiversePlace top-5 archive candidates from selected independent runs.
The largest 128 macros by area are highlighted in \textbf{\textcolor[HTML]{2D5E7E}{blue}} to make the largest macros' diverse placement modes more prominent.
A corresponding mixed-size visualization of these archive placements is provided in Appendix~\ref{app:mixed-size-diversity}.
}
\label{fig:adaptec4-top5-archive}
\end{figure*}

\begin{samepage}
\paragraph{Diversity Results.}
Table~\ref{tab:diversity} reports two diversity views.
$D_k^{\mathrm{full}}$ is the original archive diversity over complete placements, while $D_k^{\mathrm{PPO}}$ removes the WireMask greedy tail and measures only the macros still inside PPO control~\cite{shi2023wiremask}.
DiversePlace is higher than EfficientPlace for every benchmark and every archive size in both views.
In addition, Figure~\ref{fig:adaptec4-top5-archive} visualizes the archived placements on \textit{adaptec4}, providing a qualitative view of the layout modes captured by the diversity mechanism.
\par
\end{samepage}

\paragraph{Downstream Evaluation.}
We further evaluate whether the archive produced by DiversePlace transfers to downstream physical-design proxies beyond macro-placement HPWL. 
Specifically, we complete standard-cell placement with DREAMPlace~\citep{lin2021dreamplace} and evaluate routability using RUDY congestion~\citep{spindler2007rudy}, with detailed protocols and results reported in Appendix~\ref{app:mixed-size-congestion}. 
Although DiversePlace is trained primarily with HPWL, its diverse top-$k$ archive provides a more favorable candidate pool for downstream selection, creating additional opportunities to identify HPWL-competitive macro placements that align better with mixed-size placement quality or congestion profiles.
Moreover, full placement-and-routing evaluation on \textit{ariane133} demonstrates higher candidate yield and stronger post-route timing--wirelength results across three training seeds (Appendix~\ref{app:full-pr-evaluation}).
These results support the role of quality-controlled diversity as a practical interface between proxy-driven macro placement and later physical-design stages, rather than merely as an exploration mechanism.

\subsection{Ablation Studies}

\begin{figure}[H]
\centering
\begin{minipage}[t]{0.26\textwidth}
\centering
\textbf{(a)}\\[-0.2em]
\vspace{2em}
\includegraphics[width=\linewidth]{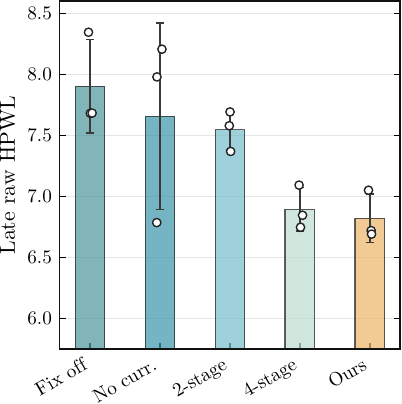}
\end{minipage}
\hfill
\begin{minipage}[t]{0.73\textwidth}
\centering
\textbf{(b)}\\[-0.2em]
\includegraphics[width=\linewidth]{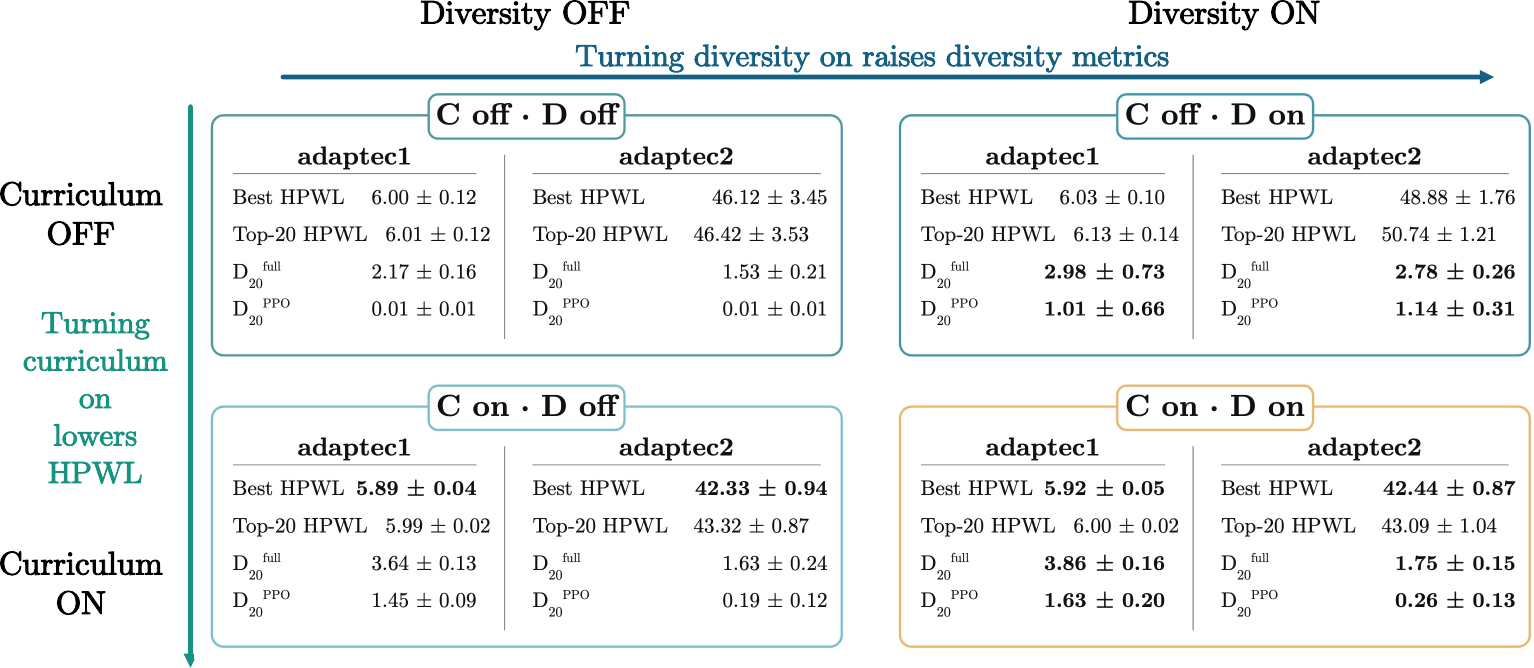}
\end{minipage}
\caption{\textbf{(a)} Late-stage raw HPWL on \textit{adaptec1}, episodes 6K--10K. \textbf{(b)} Unified $2\times2$ curriculum--diversity ablation on \textit{adaptec1} and \textit{adaptec2}. Lower HPWL and higher $D_{20}$ are better.}
\label{fig:ablation-studies}
\end{figure}

Figure~\ref{fig:ablation-studies}(a) reports late-stage raw HPWL on \textit{adaptec1} over the final 4K global episodes, after early transient exploration has largely subsided.
Removing prefix transfer gives the worst HPWL, indicating that inherited prefixes are not merely warm starts but structural priors that anchor previously optimized large macros and shape subsequent layout formation.
Among the tested curriculum variants, a finer staged expansion, corresponding to a smaller $\Delta H$ in Section~\ref{subsec:curriculum}, yields lower final-window HPWL, suggesting that a reasonably gradual curriculum can provide a smoother optimization path for late-stage placement search.
DiversePlace therefore achieves the lowest late-stage HPWL by combining a small prefix-induced stability with a finer curriculum path, avoiding the less reliable optimization induced by exposing PPO to the full macro horizon at once.

Figure~\ref{fig:ablation-studies}(b) presents a two-by-two ablation of the curriculum and diversity components in DiversePlace. 
Across both \textit{adaptec1} and \textit{adaptec2}, diversity alone substantially expands the archive but does not reliably improve HPWL, indicating that unconstrained exploration can discover different placements without providing an effective optimization path. 
By contrast, curriculum accounts for the main HPWL gains, making the long-horizon placement problem easier to optimize, but its archive remains less diverse than that of the full method. 
Adding diversity on top of curriculum consistently increases both full-placement and PPO-region diversity while preserving the HPWL advantage obtained by curriculum. 
Overall, the ablation reveals a clear division of labor: curriculum makes high-quality placement search tractable, whereas quality-guided diversity broadens the set of competitive layouts discovered within that optimized search process. Appendix~\ref{app:sensitivity-analysis} provides a more detailed analysis of key parameters.

\section{Conclusion and Limitations }

\paragraph{Conclusion.}
We presented DiversePlace, a diversity-seeking curriculum PPO framework for macro placement that searches for alternative layout modes inside a competitive HPWL region. 
By coupling a quality-gated QD objective with an expanding placement-horizon curriculum, DiversePlace improves archive diversity on all eight ISPD2005 benchmarks while also delivering stronger early efficiency and lower or closely comparable 10K-checkpoint HPWL than a recent SOTA RL baseline on all designs. 
These results suggest quality-controlled diversity as a useful design principle for RL-based physical design when downstream quality matters but is not directly measurable during training.

\paragraph{Limitations.}
DiversePlace targets early-stage macro placement under an HPWL-anchored quality gate, matching the practical setting where accurate routing or timing feedback remains too expensive to query online at every rollout. This scope lets the method preserve a fast and stable RL loop while already exposing multiple competitive layout basins, and it leaves open the integration of richer downstream surrogates when they can be incorporated at comparable cost.

\section*{Acknowledgments}
The work of Wenrui Zhou, Jiashun Liu, and Ling Pan was supported in part by National Natural Science Foundation of China 62406266. The work of Wenji Fang and Zhiyao Xie was supported in part by (RGC) CRF-YCRG C6003-24Y.

\bibliographystyle{plainnat}
\bibliography{references}

@article{markov2015progress,
  title = {Progress and Challenges in {{VLSI}} Placement Research},
  author = {Markov, Igor L. and Hu, Jin and Kim, Myung-Chul},
  journal = {Proceedings of the IEEE},
  volume = {103},
  number = {11},
  pages = {1985--2003},
  year = {2015},
  doi = {10.1109/JPROC.2015.2478963}
}

@article{kirkpatrick1983optimization,
  title = {Optimization by Simulated Annealing},
  author = {Kirkpatrick, S. and Gelatt, C. D. and Vecchi, M. P.},
  journal = {Science},
  volume = {220},
  number = {4598},
  pages = {671--680},
  year = {1983},
  doi = {10.1126/science.220.4598.671}
}

@inproceedings{kahng2021advancing,
  title     = {Advancing Placement},
  author    = {Kahng, Andrew B.},
  booktitle = {Proceedings of the 2021 International Symposium on Physical Design},
  series    = {ISPD '21},
  pages     = {15--22},
  year      = {2021},
  publisher = {Association for Computing Machinery},
  doi       = {10.1145/3439706.3446884}
}

@inproceedings{gao2023scaling,
  title = {Scaling Laws for Reward Model Overoptimization},
  author = {Gao, Leo and Schulman, John and Hilton, Jacob},
  booktitle = {Proceedings of the 40th International Conference on Machine Learning},
  pages = {10835--10866},
  year = {2023},
  editor = {Krause, Andreas and Brunskill, Emma and Cho, Kyunghyun and Engelhardt, Barbara and Sabato, Sivan and Scarlett, Jonathan},
  volume = {202},
  series = {Proceedings of Machine Learning Research},
  publisher = {PMLR}
}

@inproceedings{agnesina2023autodmp,
  title     = {{AutoDMP}: Automated {DREAMPlace}-based Macro Placement},
  author    = {Agnesina, Anthony and Rajvanshi, Puranjay and Yang, Tian and Pradipta, Geraldo and Jiao, Austin and Keller, Ben and Khailany, Brucek and Ren, Haoxing},
  booktitle = {Proceedings of the 2023 International Symposium on Physical Design},
  series    = {ISPD '23},
  pages     = {149--157},
  year      = {2023},
  publisher = {Association for Computing Machinery},
  doi       = {10.1145/3569052.3578923}
}

@inproceedings{jain2023biologicalsequencedesigngflownets,
  title = {Biological Sequence Design with {GFlowNets}},
  author = {Jain, Moksh and Bengio, Emmanuel and Hernandez-Garcia, Alex and Rector-Brooks, Jarrid and Dossou, Bonaventure F. P. and Ekbote, Chanakya Ajit and Fu, Jie and Zhang, Tianyu and Kilgour, Michael and Zhang, Dinghuai and Simine, Lena and Das, Payel and Bengio, Yoshua},
  booktitle = {Proceedings of the 39th International Conference on Machine Learning},
  series = {Proceedings of Machine Learning Research},
  volume = {162},
  pages = {9786--9801},
  year = {2022},
  publisher = {PMLR},
  url = {https://proceedings.mlr.press/v162/jain22a.html}
}

@article{murata1996sequencepair,
  title = {VLSI Module Placement Based on Rectangle-Packing by the Sequence-Pair},
  author = {Murata, Hiroshi and Fujiyoshi, Kunihiro and Nakatake, Shinji and Kajitani, Yoshiyuki},
  journal = {IEEE Transactions on Computer-Aided Design of Integrated Circuits and Systems},
  volume = {15},
  number = {12},
  pages = {1518--1524},
  year = {1996},
  doi = {10.1109/43.552084}
}

@inproceedings{chang2000btree,
  title = {{{B}$^\ast$}-Trees: A New Representation for Non-Slicing Floorplans},
  author = {Chang, Yun-Chih and Chang, Yao-Wen and Wu, Guang-Ming and Wu, Shu-Wei},
  booktitle = {Proceedings of the 37th Design Automation Conference},
  pages = {458--463},
  year = {2000},
  doi = {10.1145/337292.337541}
}

@article{cheng2019replace,
  title = {{{RePlAce}}: Advancing Solution Quality and Routability Validation in Global Placement},
  author = {Cheng, Chung-Kuan and Kahng, Andrew B. and Kang, Ilgweon and Wang, Lutong},
  journal = {IEEE Transactions on Computer-Aided Design of Integrated Circuits and Systems},
  volume = {38},
  number = {9},
  pages = {1717--1730},
  year = {2019},
  doi = {10.1109/TCAD.2018.2859220}
}

@article{lin2021dreamplace,
  title = {{{DREAMPlace}}: Deep Learning Toolkit-Enabled {{GPU}} Acceleration for Modern {{VLSI}} Placement},
  author = {Lin, Yibo and Jiang, Zixuan and Gu, Jiaqi and Li, Wuxi and Dhar, Shounak and Ren, Haoxing and Khailany, Brucek and Pan, David Z.},
  journal = {IEEE Transactions on Computer-Aided Design of Integrated Circuits and Systems},
  volume = {40},
  number = {4},
  pages = {748--761},
  year = {2021},
  doi = {10.1109/TCAD.2020.3003843}
}

@inproceedings{liu2022floorplanning,
  title = {Floorplanning with Graph Attention},
  author = {Liu, Yiting and Ju, Ziyi and Li, Zhengming and Dong, Mingzhi and Zhou, Hai and Wang, Jia and Yang, Fan and Zeng, Xuan and Shang, Li},
  booktitle = {Proceedings of the 59th ACM/IEEE Design Automation Conference},
  pages = {1303--1308},
  year = {2022},
  doi = {10.1145/3489517.3530484}
}

@article{liu2023graphplanner,
  title = {{{GraphPlanner}}: Floorplanning with Graph Neural Network},
  author = {Liu, Yiting and Ju, Ziyi and Li, Zhengming and Dong, Mingzhi and Zhou, Hai and Wang, Jia and Yang, Fan and Zeng, Xuan and Li, Shang},
  journal = {ACM Transactions on Design Automation of Electronic Systems},
  volume = {28},
  number = {2},
  pages = {21:1--21:24},
  year = {2023},
  doi = {10.1145/3555804}
}

@article{lu2024ganplace,
  title = {{{GAN-Place}}: Advancing Open Source Placers to Commercial-Quality Using Generative Adversarial Networks and Transfer Learning},
  author = {Lu, Yi-Chen and Ren, Haoxing and Hsiao, Hao-Hsiang and Lim, Sung Kyu},
  journal = {ACM Transactions on Design Automation of Electronic Systems},
  volume = {29},
  number = {2},
  pages = {1--17},
  year = {2024},
  doi = {10.1145/3636461}
}

@inproceedings{bengio2009curriculum,
  title = {Curriculum Learning},
  author = {Bengio, Yoshua and Louradour, J{\'e}r{\^o}me and Collobert, Ronan and Weston, Jason},
  booktitle = {Proceedings of the 26th International Conference on Machine Learning},
  pages = {41--48},
  year = {2009},
  doi = {10.1145/1553374.1553380}
}

@article{mirhoseini2021graph,
  title = {A Graph Placement Methodology for Fast Chip Design},
  author = {Mirhoseini, Azalia and Goldie, Anna and Yazgan, Mustafa and Jiang, Joe Wenjie and Songhori, Ebrahim and Wang, Shen and Lee, Young-Joon and Johnson, Eric and Pathak, Omkar and Nova, Azade and Pak, Jiwoo and Tong, Andy and Srinivasa, Kavya and Hang, William and Tuncer, Emre and Le, Quoc V. and Laudon, James and Ho, Richard and Carpenter, Roger and Dean, Jeff},
  journal = {Nature},
  volume = {594},
  number = {7862},
  pages = {207--212},
  year = {2021},
  doi = {10.1038/s41586-021-03544-w}
}

@inproceedings{cheng2021deeppr,
  title = {On Joint Learning for Solving Placement and Routing in Chip Design},
  author = {Cheng, Ruoyu and Yan, Junchi},
  booktitle = {Advances in Neural Information Processing Systems},
  volume = {34},
  pages = {16508--16519},
  year = {2021}
}

@inproceedings{cheng2022prnet,
  title = {The Policy-Gradient Placement and Generative Routing Neural Networks for Chip Design},
  author = {Cheng, Ruoyu and Lyu, Xianglong and Li, Yang and Ye, Junjie and Hao, Jianye and Yan, Junchi},
  booktitle = {Advances in Neural Information Processing Systems},
  volume = {35},
  year = {2022}
}

@article{lai2022maskplace,
  title = {MaskPlace: Fast Chip Placement via Reinforced Visual Representation Learning},
  author = {Lai, Yao and Mu, Yao and Luo, Ping},
  journal = {Advances in Neural Information Processing Systems},
  volume = {35},
  pages = {24019--24030},
  year = {2022}
}

@inproceedings{lai2023chipformer,
  title = {{{ChiPFormer}}: Transferable Chip Placement via Offline Decision Transformer},
  author = {Lai, Yao and Liu, Jinxin and Tang, Zhentao and Wang, Bin and Hao, Jianye and Luo, Ping},
  booktitle = {Proceedings of the 40th International Conference on Machine Learning},
  series = {Proceedings of Machine Learning Research},
  volume = {202},
  pages = {18346--18364},
  year = {2023}
}

@inproceedings{shi2023wiremask,
  title = {Macro Placement by Wire-Mask-Guided Black-Box Optimization},
  author = {Shi, Yunqi and Xue, Ke and Lei, Song and Qian, Chao},
  booktitle = {Advances in Neural Information Processing Systems},
  volume = {36},
  year = {2023}
}

@inproceedings{geng2024efficientplace,
  title = {Reinforcement Learning within Tree Search for Fast Macro Placement},
  author = {Geng, Zijie and Wang, Jie and Liu, Ziyan and Xu, Siyuan and Tang, Zhentao and Yuan, Mingxuan and Hao, Jianye and Zhang, Yongdong and Wu, Feng},
  booktitle = {Proceedings of the 41st International Conference on Machine Learning},
  series = {Proceedings of Machine Learning Research},
  volume = {235},
  pages = {15402--15417},
  year = {2024}
}

@inproceedings{spindler2007rudy,
  title = {Fast and Accurate Routing Demand Estimation for Efficient Routability-Driven Placement},
  author = {Spindler, Peter and Johannes, Frank M.},
  booktitle = {Design, Automation and Test in Europe Conference and Exhibition},
  pages = {1--6},
  year = {2007},
  organization = {IEEE},
  doi = {10.1109/DATE.2007.364463}
}

@inproceedings{nam2005ispd,
  title = {The {{ISPD}}2005 Placement Contest and Benchmark Suite},
  author = {Nam, Gi-Joon and Alpert, Charles J. and Villarrubia, Paul and Winter, Bruce and Yildiz, Mehmet Can},
  booktitle = {Proceedings of the 2005 International Symposium on Physical Design},
  pages = {216--220},
  year = {2005},
  doi = {10.1145/1055137.1055182}
}

@inproceedings{lee2025diffusion,
  title = {Chip Placement with Diffusion Models},
  author = {Lee, Vint and Nguyen, Minh and Elzeiny, Leena and Deng, Chun and Abbeel, Pieter and Wawrzynek, John},
  booktitle = {Proceedings of the 42nd International Conference on Machine Learning},
  series = {Proceedings of Machine Learning Research},
  volume = {267},
  pages = {33499--33514},
  year = {2025},
  url = {https://proceedings.mlr.press/v267/lee25y.html}
}

@inproceedings{eysenbach2019diversity,
  title = {Diversity Is All You Need: Learning Skills without a Reward Function},
  author = {Eysenbach, Benjamin and Gupta, Abhishek and Ibarz, Julian and Levine, Sergey},
  booktitle = {International Conference on Learning Representations},
  year = {2019},
  url = {https://openreview.net/forum?id=SJx63jRqFm}
}

@article{mouret2015illuminating,
  title = {Illuminating Search Spaces by Mapping Elites},
  author = {Mouret, Jean-Baptiste and Clune, Jeff},
  journal = {arXiv preprint arXiv:1504.04909},
  year = {2015},
  doi = {10.48550/arXiv.1504.04909},
  url = {https://arxiv.org/abs/1504.04909}
}

@article{pugh2016qualitydiversity,
  title = {Quality Diversity: A New Frontier for Evolutionary Computation},
  author = {Pugh, Justin K. and Soros, Lisa B. and Stanley, Kenneth O.},
  journal = {Frontiers in Robotics and AI},
  volume = {3},
  pages = {40},
  year = {2016},
  doi = {10.3389/frobt.2016.00040},
  url = {https://www.frontiersin.org/articles/10.3389/frobt.2016.00040/full}
}

@article{schulman2017ppo,
  title = {Proximal Policy Optimization Algorithms},
  author = {Schulman, John and Wolski, Filip and Dhariwal, Prafulla and Radford, Alec and Klimov, Oleg},
  journal = {arXiv preprint arXiv:1707.06347},
  year = {2017},
  doi = {10.48550/arXiv.1707.06347},
  url = {https://arxiv.org/abs/1707.06347}
}

@inproceedings{xue2024macroregulator,
 author = {Xue, Ke and Chen, Ruo-Tong and Lin, Xi and Shi, Yunqi and Kai, Shixiong and Xu, Siyuan and Qian, Chao},
 booktitle = {Advances in Neural Information Processing Systems},
 doi = {10.52202/079017-4462},
 editor = {A. Globerson and L. Mackey and D. Belgrave and A. Fan and U. Paquet and J. Tomczak and C. Zhang},
 pages = {140565--140588},
 publisher = {Curran Associates, Inc.},
 title = {Reinforcement Learning Policy as Macro Regulator Rather than Macro Placer},
 url = {https://proceedings.neurips.cc/paper_files/paper/2024/file/fe224a60b878e79d5b3d79d7f113f76b-Paper-Conference.pdf},
 volume = {37},
 year = {2024}
}

@inproceedings{gao2026expertise,
 author = {Gao, Chengrui and Shi, Yunqi and Xue, Ke and Chen, Ruo-Tong and Xu, Siyuan and Yuan, Mingxuan and Qian, Chao and Zhou, Zhi-Hua},
 booktitle = {International Conference on Learning Representations},
 editor = {C. Vondrick and B. Hariharan and C. Raffel and L. Pinto and D. Yang and A. Faust},
 pages = {125611--125634},
 title = {Expertise Can Be Helpful for Reinforcement Learning-based Macro Placement},
 url = {https://proceedings.iclr.cc/paper_files/paper/2026/file/cbc1ad2066f0afebbcea930c5688fc1f-Paper-Conference.pdf},
 volume = {2026},
 year = {2026}
}

\clearpage
\appendix
\setcounter{figure}{0}
\setcounter{table}{0}
\makeatletter
\@addtoreset{figure}{section}
\@addtoreset{table}{section}
\makeatother
\renewcommand{\thefigure}{\thesection\arabic{figure}}
\renewcommand{\theHfigure}{\thesection.\arabic{figure}}
\renewcommand{\thetable}{\thesection\arabic{table}}
\renewcommand{\theHtable}{\thesection.\arabic{table}}
\begingroup
\raggedbottom
\setlength{\floatsep}{6pt plus 2pt minus 2pt}
\setlength{\textfloatsep}{8pt plus 2pt minus 2pt}
\setlength{\intextsep}{6pt plus 2pt minus 2pt}
\setlength{\parskip}{2pt}
\makeatletter
\setlength{\@neuripsabovecaptionskip}{4pt}
\makeatother
\setlength{\abovecaptionskip}{4pt}

\section{Dataset Description}
\label{app:dataset-description}

We evaluate on the eight public ISPD2005 placement-contest circuits, \textit{adaptec1--4} and \textit{bigblue1--4}~\citep{nam2005ispd} following prior work~\citep{lai2022maskplace,geng2024efficientplace}.
We treat each circuit as a separate placement instance with its own canvas, fixed terminals, macro geometry, standard-cell population, netlist connectivity, and utilization profile.
Table~\ref{tab:appendix-dataset-description} summarizes the public circuit statistics reported for the ISPD2005 instances used in our experiments.

\begin{table*}[!htbp]
\centering
\scriptsize
\setlength{\tabcolsep}{3.0pt}
\caption{
Statistics of the ISPD2005 benchmark circuits used in our evaluation.
}
\label{tab:appendix-dataset-description}
\resizebox{\textwidth}{!}{%
\begin{tabular}{@{}cccccccc@{}}
\toprule
Circuit & Macros & Macros placed & Hard macros & Standard cells & Nets & Pins & Area util. (\%) \\
\midrule
\textit{adaptec1} & 543 & 543 & 63 & 210904 & 221142 & 944063 & 55.62 \\
\textit{adaptec2} & 566 & 566 & 159 & 254457 & 266009 & 1069482 & 74.46 \\
\textit{adaptec3} & 723 & 723 & 201 & 450927 & 466758 & 1875039 & 61.51 \\
\textit{adaptec4} & 1329 & 1329 & 92 & 494716 & 515951 & 1912420 & 48.62 \\
\textit{bigblue1} & 560 & 560 & 32 & 277604 & 284479 & 1144691 & 31.58 \\
\textit{bigblue2} & 23084 & 256 & 52 & 534782 & 577235 & 2122282 & 32.43 \\
\textit{bigblue3} & 1293 & 1293 & 138 & 1095519 & 1123170 & 3833218 & 66.81 \\
\textit{bigblue4} & 8170 & 1024 & 52 & 2169183 & 2229886 & 8900078 & 35.68 \\
\bottomrule
\end{tabular}%
}
\end{table*}

Although these circuits originate from the ISPD2005 challenge, \textbf{they should not be interpreted as sharing a single repeated layout pattern.}
Their physical and netlist formations vary sharply.
The total macro count ranges from 543 to 23084, the standard-cell count ranges from roughly $2.1\times 10^5$ to $2.2\times 10^6$, the pin count spans almost an order of magnitude, and utilization ranges from 31.58\% to 74.46\%.
These differences change the placement search landscape, including available whitespace, macro blockage structure, fixed-terminal pressure, net bounding-box behavior, and the interaction between macro positions and later standard-cell placement.
Therefore, strong results across all eight circuits do not merely indicate that a method fits one benchmark source.
They provide evidence that the method remains effective across heterogeneous circuit formations, supporting the robustness and generalizability of DiversePlace under substantially different placement instances.

\FloatBarrier
\section{Illustration of HPWL Update}
\label{app:hpwl-update-illustration}

In this section, we illustrate how the addition of a macro updates the bounding boxes used for HPWL calculation in Figure~\ref{fig:hpwl-update-illustration}.
Specifically, when a macro is added to a partial placement, each affected net recomputes its bounding box from the extrema of its placed pins, with the HPWL increment equal to the resulting change in half-perimeter.

\begin{figure*}[!htbp]
\centering
\begin{minipage}[t]{0.43\textwidth}
  \centering
  \includegraphics[width=\linewidth]{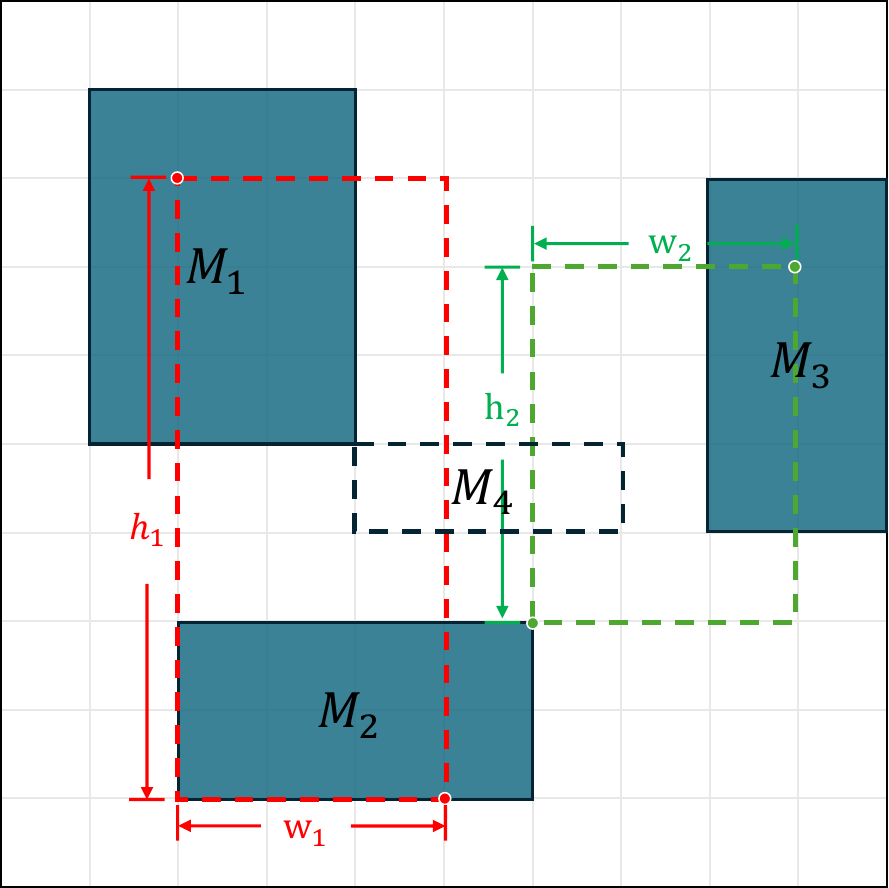}

  \vspace{0.4ex}
  \small\textbf{(a)} Before placing $M_4$.
\end{minipage}\hfill
\begin{minipage}[t]{0.43\textwidth}
  \centering
  \includegraphics[width=\linewidth]{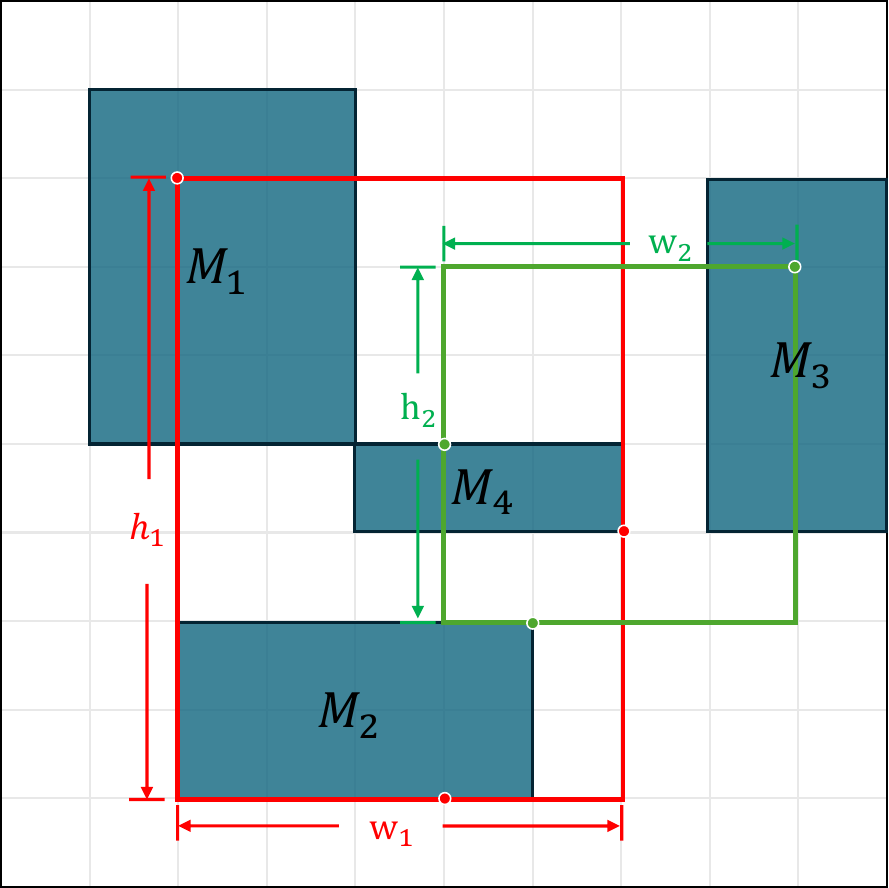}

  \vspace{0.4ex}
  \small\textbf{(b)} After placing $M_4$.
\end{minipage}
\caption{\textbf{Adding $M_4$ changes affected nets' HPWL bounding boxes.}
In partial placement (a), dashed outlines mark $M_4$'s candidate location (black) and the current net bounding boxes (red and green).
In (b), $M_4$ enlarges the affected pin sets; new extrema expand or shift the recomputed boxes, changing each net's half-perimeter contribution.
The two affected nets have total HPWL $h_1 + w_1 + h_2 + w_2$, with $(h_i, w_i)$ denoting the $i$-th net bounding box's height and width.
}
\label{fig:hpwl-update-illustration}
\end{figure*}

\FloatBarrier
\section{Additional Experiment Results and Figures}
\label{app:additional-experiment-figures}

Figure~\ref{fig:hpwl-trajectory} reports the full cumulative-best HPWL trajectories behind the aggregate results in Section~\ref{sec:experiments}.
The curves expose both final placement quality and convergence behavior across the complete training budget.
Table~\ref{tab:appendix-current-progress-matrix} complements Table~\ref{tab:hpwl-overall} with best-observed HPWL at 1K, 2K, and 10K episodes and top-$k$ archive statistics across all eight benchmarks.
Its 10K archive HPWL and full-placement diversity values match Table~\ref{tab:diversity}.

\begin{figure}[!htbp]
\centering
\includegraphics[width=0.98\textwidth]{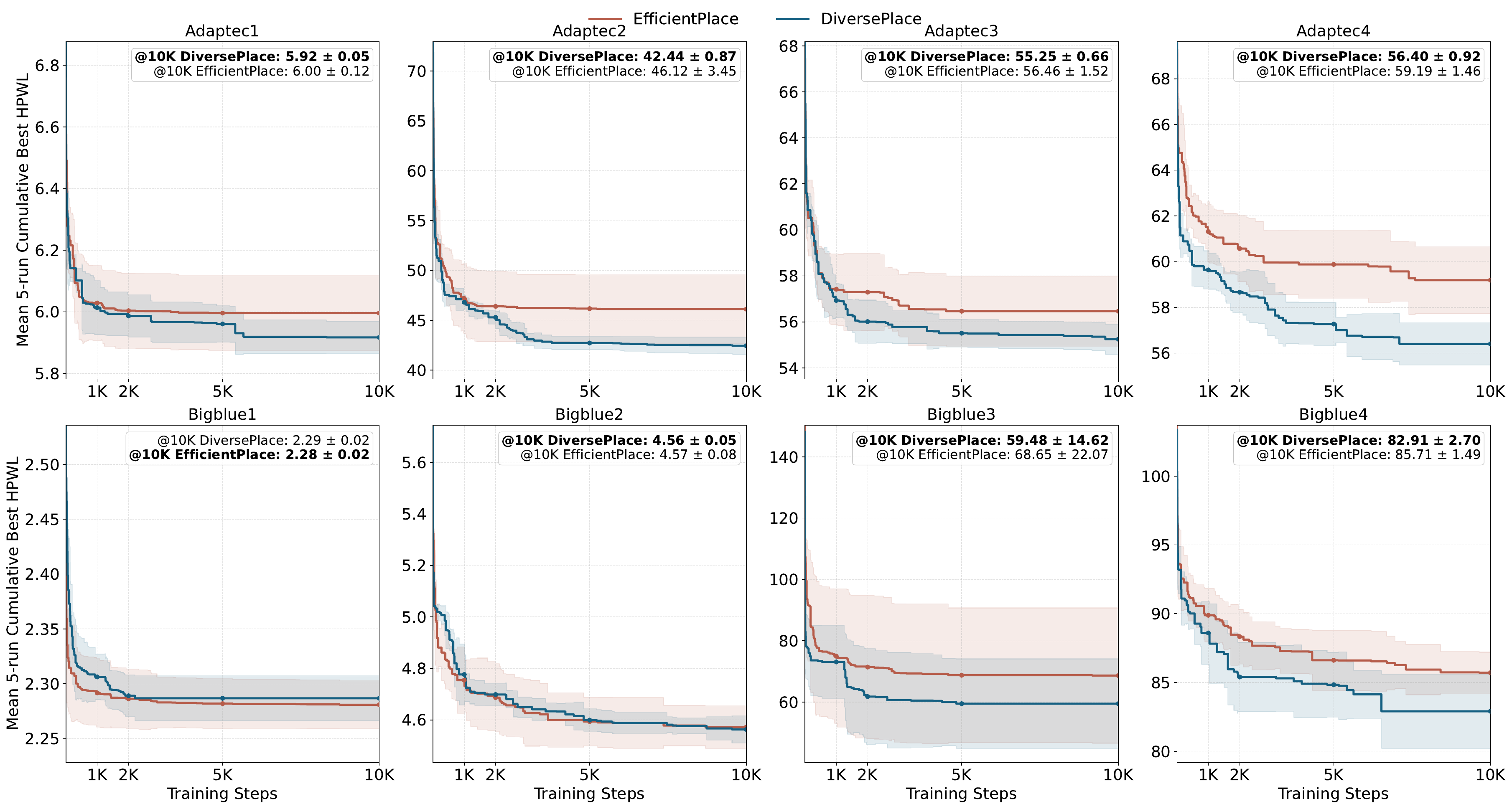}
\caption{Cumulative-best HPWL trajectories of DiversePlace and EfficientPlace reconstructed from the original run logs. Lower is better.}
\label{fig:hpwl-trajectory}
\end{figure}

\begin{table}[!htbp]
\centering
\tiny
\setlength{\tabcolsep}{1.45pt}
\renewcommand{\arraystretch}{0.95}
{\fontfamily{cmr}\selectfont
\caption{Summary of best-observed HPWL and top-$k$ archive statistics.
Each dataset has two rows: DiversePlace (D) and EfficientPlace (E). 
B denotes best-observed HPWL and follows Table~\ref{tab:hpwl-overall}.
T5/T10/T20 denote mean HPWL within the top-5/top-10/top-20 archives; D5/D10/D20 denote full-placement diversity at 10K.
The 10K archive HPWL and diversity values match Table~\ref{tab:diversity}.
Values are rounded to two decimals. 
Lower HPWL and higher diversity win; winners after rounding are shown in \textbf{bold}.}
\label{tab:appendix-current-progress-matrix}
\resizebox{\textwidth}{!}{%
\begin{tabular}{@{}cc*{15}{c}@{}}
\toprule
Dataset & Method & \multicolumn{4}{c}{1K HPWL} & \multicolumn{4}{c}{2K HPWL} & \multicolumn{4}{c}{10K HPWL} & \multicolumn{3}{c}{10K Diversity} \\
\cmidrule(lr){3-6}\cmidrule(lr){7-10}\cmidrule(lr){11-14}\cmidrule(lr){15-17}
 &  & B & T5 & T10 & T20 & B & T5 & T10 & T20 & B & T5 & T10 & T20 & D5 & D10 & D20 \\
\midrule
Adaptec1 & D & \textbf{6.01} & 6.08 & 6.11 & 6.16 & \textbf{5.99} & 6.04 & 6.06 & 6.09 & \textbf{5.91} & \textbf{5.96} & \textbf{5.98} & \textbf{6.00} & \textbf{3.98} & \textbf{3.89} & \textbf{3.86} \\
 & E & 6.03 & \textbf{6.05} & \textbf{6.05} & \textbf{6.07} & 6.00 & \textbf{6.02} & \textbf{6.02} & \textbf{6.03} & 6.00 & 6.00 & 6.00 & 6.01 & 2.13 & 2.17 & 2.17 \\
Adaptec2 & D & \textbf{46.81} & 47.92 & 48.60 & 49.40 & \textbf{44.59} & \textbf{45.18} & \textbf{45.50} & \textbf{45.97} & \textbf{42.44} & \textbf{42.77} & \textbf{42.92} & \textbf{43.08} & \textbf{1.83} & \textbf{1.77} & \textbf{1.75} \\
 & E & 47.19 & \textbf{47.66} & \textbf{47.94} & \textbf{48.24} & 46.40 & 46.53 & 46.66 & 46.81 & 46.12 & 46.24 & 46.33 & 46.42 & 1.48 & 1.51 & 1.53 \\
Adaptec3 & D & \textbf{56.93} & \textbf{57.47} & \textbf{57.95} & 58.62 & \textbf{55.48} & \textbf{56.14} & \textbf{56.43} & \textbf{56.80} & \textbf{55.25} & \textbf{55.82} & \textbf{56.01} & \textbf{56.25} & \textbf{4.20} & \textbf{4.28} & \textbf{4.34} \\
 & E & 57.42 & 57.84 & 58.11 & \textbf{58.41} & 57.29 & 57.53 & 57.73 & 57.97 & 56.46 & 56.87 & 57.05 & 57.24 & 3.41 & 3.62 & 3.68 \\
Adaptec4 & D & \textbf{59.64} & \textbf{60.67} & \textbf{61.16} & \textbf{61.74} & \textbf{58.66} & \textbf{59.10} & \textbf{59.44} & \textbf{59.94} & \textbf{56.40} & \textbf{57.16} & \textbf{57.58} & \textbf{58.00} & \textbf{10.89} & \textbf{11.42} & \textbf{11.69} \\
 & E & 61.32 & 62.07 & 62.42 & 62.85 & 60.57 & 61.05 & 61.35 & 61.76 & 59.19 & 59.88 & 60.21 & 60.55 & 8.56 & 8.75 & 8.82 \\
Bigblue1 & D & 2.31 & 2.31 & 2.32 & 2.32 & 2.29 & 2.29 & \textbf{2.29} & 2.30 & 2.29 & 2.29 & 2.29 & 2.30 & \textbf{3.80} & \textbf{3.82} & \textbf{3.84} \\
 & E & \textbf{2.29} & \textbf{2.30} & \textbf{2.31} & \textbf{2.31} & 2.29 & 2.29 & 2.30 & 2.30 & \textbf{2.28} & 2.29 & 2.29 & \textbf{2.29} & 2.79 & 2.76 & 2.71 \\
Bigblue2 & D & 4.78 & 4.86 & 4.91 & 4.96 & 4.70 & 4.73 & 4.76 & 4.80 & \textbf{4.56} & 4.61 & 4.63 & 4.66 & \textbf{2.28} & \textbf{2.20} & \textbf{2.49} \\
 & E & \textbf{4.75} & \textbf{4.80} & \textbf{4.82} & \textbf{4.86} & \textbf{4.69} & 4.73 & \textbf{4.75} & \textbf{4.77} & 4.57 & 4.61 & \textbf{4.62} & \textbf{4.64} & 0.62 & 0.62 & 0.67 \\
Bigblue3 & D & \textbf{73.10} & 82.14 & 86.78 & 93.63 & \textbf{61.81} & \textbf{64.58} & \textbf{65.80} & \textbf{67.29} & \textbf{59.48} & \textbf{60.92} & \textbf{61.77} & \textbf{62.75} & \textbf{4.57} & \textbf{4.66} & \textbf{4.65} \\
 & E & 74.96 & \textbf{76.57} & \textbf{77.87} & \textbf{79.22} & 71.43 & 72.70 & 73.41 & 74.40 & 68.65 & 69.75 & 70.32 & 70.98 & 3.37 & 3.38 & 3.47 \\
Bigblue4 & D & \textbf{88.60} & \textbf{90.14} & \textbf{91.09} & \textbf{92.17} & \textbf{85.40} & \textbf{87.45} & \textbf{88.72} & \textbf{90.01} & \textbf{82.91} & \textbf{85.05} & \textbf{86.03} & \textbf{86.89} & \textbf{6.13} & \textbf{6.24} & \textbf{6.26} \\
 & E & 89.88 & 91.36 & 92.21 & 93.18 & 88.33 & 89.41 & 90.08 & 90.91 & 85.71 & 86.79 & 87.28 & 87.90 & 4.42 & 4.27 & 4.28 \\
\bottomrule
\end{tabular}%
}
}
\end{table}

\FloatBarrier
\section{Mixed-Size Placement and Congestion Evaluation}
\label{app:mixed-size-congestion}

\paragraph{RUDY overflow metric.}
Rectangular uniform wire density (RUDY) is first computed as a two-dimensional map over the final layout.
Each bin in this map estimates the local routing utilization induced by net bounding boxes; values above one indicate demand beyond the normalized routing capacity of that bin.
To summarize the whole layout with one scalar, we report the normalized overflow of this map:
\begin{equation}
\mathrm{RUDYOverflow}(P)=
\left\{
\begin{array}{ll}
\frac{\sum_{b \in B}\max(U_b-1,0)}{\sum_{b \in B}U_b}, & \sum_{b \in B}U_b>0,\\
0, & \sum_{b \in B}U_b=0,
\end{array}
\right.
\end{equation}
where $P$ is the evaluated placement, $B$ is the routing-bin grid, and $U_b=\max(U^h_b,U^v_b)$ is the larger of the horizontal and vertical capacity-normalized RUDY utilizations in bin $b$.
Thus, the scalar is not the mean RUDY of the image.
It is the total amount by which congested bins exceed the capacity threshold, normalized by total estimated routing demand.
Algorithm~\ref{alg:rudy-overflow} summarizes the computation.
\vspace{3pt}

\begin{breakablealgorithm}
\caption{RUDY overflow for one completed placement}
\label{alg:rudy-overflow}
\begin{algorithmic}[1]
\Require Completed mixed-size placement $P$, netlist $N$, routing-bin grid $B$, directional capacities $c_h,c_v$
\Ensure Scalar RUDY overflow
\State Initialize horizontal and vertical demand maps $D^h_b,D^v_b \gets 0$ for all $b \in B$
\ForAll{nets $n \in N$}
    \State Collect final pin coordinates of $n$ from $P$
    \If{$n$ has fewer than two placed pins}
        \State \textbf{continue}
    \EndIf
    \State Compute the bounding box of $n$ with width $W_n$ and height $H_n$
    \ForAll{bins $b$ overlapped by the bounding box of $n$}
        \State Compute overlap area $A_{n,b}$ between bin $b$ and the bounding box
        \State $D^h_b \gets D^h_b+\omega_n A_{n,b}/(H_n+\epsilon)$
        \State $D^v_b \gets D^v_b+\omega_n A_{n,b}/(W_n+\epsilon)$
    \EndFor
\EndFor
\ForAll{bins $b \in B$}
    \State $U^h_b \gets D^h_b/(\mathrm{area}(b)c_h)$; \quad $U^v_b \gets D^v_b/(\mathrm{area}(b)c_v)$
    \State $U_b \gets \max(U^h_b,U^v_b)$
\EndFor
\If{$\sum_{b \in B}U_b > 0$}
    \State \Return $\sum_{b \in B}\max(U_b-1,0) / \sum_{b \in B}U_b$
\Else
    \State \Return $0$
\EndIf
\end{algorithmic}
\end{breakablealgorithm}

\textbf{The RUDY signal is only available after mixed-size placement.}
During macro-placement training, the standard cells do not yet have final physical locations, so the complete post-refinement pin coordinates required by the RUDY map are unavailable.
We therefore use RUDY only as a downstream evaluation and archive-selection signal after DREAMPlace finishes mixed-size placement; it is not used as a training-time reward.
All EfficientPlace and DiversePlace candidates use the same routing-bin resolution for a given circuit during RUDY evaluation.
Table~\ref{tab:appendix-rudy-runtime} reports the average evaluation time per completed candidate in our cached runs.

The combined row sums phase-1 DREAMPlace cell placement, phase-2 DREAMPlace mixed-size placement, and RUDY evaluation; it excludes macro-placement training and archive generation.
These costs are for evaluating a single completed layout.
If RUDY were incorporated directly into RL training as a reward or selection signal, each episode or candidate requiring such feedback would need to run the downstream placement-and-congestion evaluation before the signal becomes available.
This would make training prohibitively slow, so we treat RUDY as a time-consuming downstream metric rather than a training-time signal.

\begin{table*}[!htbp]
\centering
\scriptsize
\setlength{\tabcolsep}{3.0pt}
\begin{tabular}{@{}lcccccccc@{}}
\toprule
Circuit & adaptec1 & adaptec2 & adaptec3 & adaptec4 & bigblue1 & bigblue2 & bigblue3 & bigblue4 \\
\midrule
RUDY bins & $512^2$ & $1024^2$ & $1024^2$ & $1024^2$ & $512^2$ & $1024^2$ & $2048^2$ & $2048^2$ \\
RUDY only (min.) & 1.85 & 6.05 & 8.00 & 7.05 & 2.26 & 4.76 & 37.75 & 100.71 \\
DREAMPlace+RUDY (min.) & 2.85 & 7.49 & 11.12 & 10.89 & 3.64 & 7.58 & 45.26 & 118.75 \\
\bottomrule
\end{tabular}
\caption{
Average downstream evaluation cost per completed mixed-size placement candidate.
`RUDY only' measures congestion evaluation after the final phase-2 placement is already available.
`DREAMPlace+RUDY' sums phase-1 DREAMPlace placement, phase-2 DREAMPlace mixed-size placement, and RUDY evaluation.
Both EfficientPlace and DiversePlace use the same routing-bin resolution for each benchmark during RUDY evaluation.}
\label{tab:appendix-rudy-runtime}
\end{table*}

\paragraph{Mixed-size placement.}
We follow a standard two-stage mixed-size placement practice used by recent RL-based placement research~\citep{geng2024efficientplace}.
Given macro positions produced by a macro placer, DREAMPlace first performs standard-cell placement with macros fixed and without legalization.
The resulting placement initializes a second DREAMPlace mixed-size placement run with global placement, legalization, and detailed placement enabled.
The DREAMPlace row in Table~\ref{tab:appendix-mixed-size} is reported accordingly~\citep{shi2023wiremask,geng2024efficientplace}.
For each seed and each method, we first take the top-$k$ macro archive ranked by macro-placement HPWL, run the same DREAMPlace mixed-size flow for every candidate in this archive, and then report the best downstream mixed-size HPWL within that archive.

\begin{table*}[!htbp]
\centering
\scriptsize
\setlength{\tabcolsep}{1.2pt}
\caption{
Mixed-size placement HPWL comparison on ISPD2005 circuits.
Values are final DREAMPlace mixed-size HPWL in $\times 10^7$ units; lower is better.
The standalone DREAMPlace results follow prior reported values~\citep{geng2024efficientplace}.
For EfficientPlace and DiversePlace, we evaluate the corresponding macro placements from Table~\ref{tab:hpwl-overall} through the same downstream DREAMPlace mixed-size flow.
Results are averaged over five predetermined independent runs with seeds 0--4 and reported as mean $\pm$ sample standard deviation.
}
\label{tab:appendix-mixed-size}
\resizebox{\textwidth}{!}{%
\begin{tabular}{@{}C{3.0cm}*{8}{C{1.45cm}}@{}}
\toprule
Method & adaptec1 & adaptec2 & adaptec3 & adaptec4 & bigblue1 & bigblue2 & bigblue3 & bigblue4 \\
\midrule
DREAMPlace & 11.10$\pm$1.31 & 13.84$\pm$1.74 & 17.03$\pm$0.99 & 24.37$\pm$1.13 & 10.06$\pm$0.28 & -- & 36.51$\pm$0.56 & 175.86$\pm$2.23 \\
Top-5 EfficientPlace+DREAMPlace & 10.42$\pm$0.35 & 11.24$\pm$1.49 & 24.20$\pm$2.64 & \textbf{22.66$\pm$0.56} & 10.41$\pm$0.32 & 14.79$\pm$0.13 & 50.07$\pm$1.61 & 105.65$\pm$4.95 \\
Top-5 DiversePlace+DREAMPlace & \textbf{9.53$\pm$0.44} & \textbf{10.82$\pm$0.31} & \textbf{22.15$\pm$1.68} & 23.00$\pm$1.56 & 10.41$\pm$0.23 & \textbf{14.68$\pm$0.17} & \textbf{41.45$\pm$4.50} & \textbf{105.30$\pm$2.78} \\
Top-10 EfficientPlace+DREAMPlace & 10.19$\pm$0.41 & 11.16$\pm$1.50 & 23.41$\pm$2.23 & 22.63$\pm$0.57 & \textbf{10.31$\pm$0.37} & 14.77$\pm$0.12 & 49.83$\pm$1.80 & 105.38$\pm$5.16 \\
Top-10 DiversePlace+DREAMPlace & \textbf{9.48$\pm$0.38} & \textbf{10.65$\pm$0.48} & \textbf{21.94$\pm$1.47} & \textbf{22.61$\pm$1.70} & 10.36$\pm$0.18 & \textbf{14.66$\pm$0.14} & \textbf{39.82$\pm$3.60} & \textbf{105.12$\pm$2.83} \\
\bottomrule
\end{tabular}%
}
\end{table*}

Table~\ref{tab:appendix-mixed-size} shows that the benefits of a diverse macro-placement archive carry over to mixed-size placement.
Across both top-5 and top-10 archives, DiversePlace yields better DREAMPlace-refined HPWL than EfficientPlace on most circuits, indicating a greater chance of finding a candidate that better survives standard-cell placement and refinement.
Downstream-aware selection from the archive can recover solutions that would be missed by a single proxy-optimal placement.
Thus, quality-guided diversity supports exploration and provides a practical interface between macro optimization and subsequent physical-design stages.
Figure~\ref{fig:appendix-mixed-size-placements} visualizes mixed-size placements for every dataset.

\begin{figure*}[!htbp]
\centering
\includegraphics[width=\textwidth]{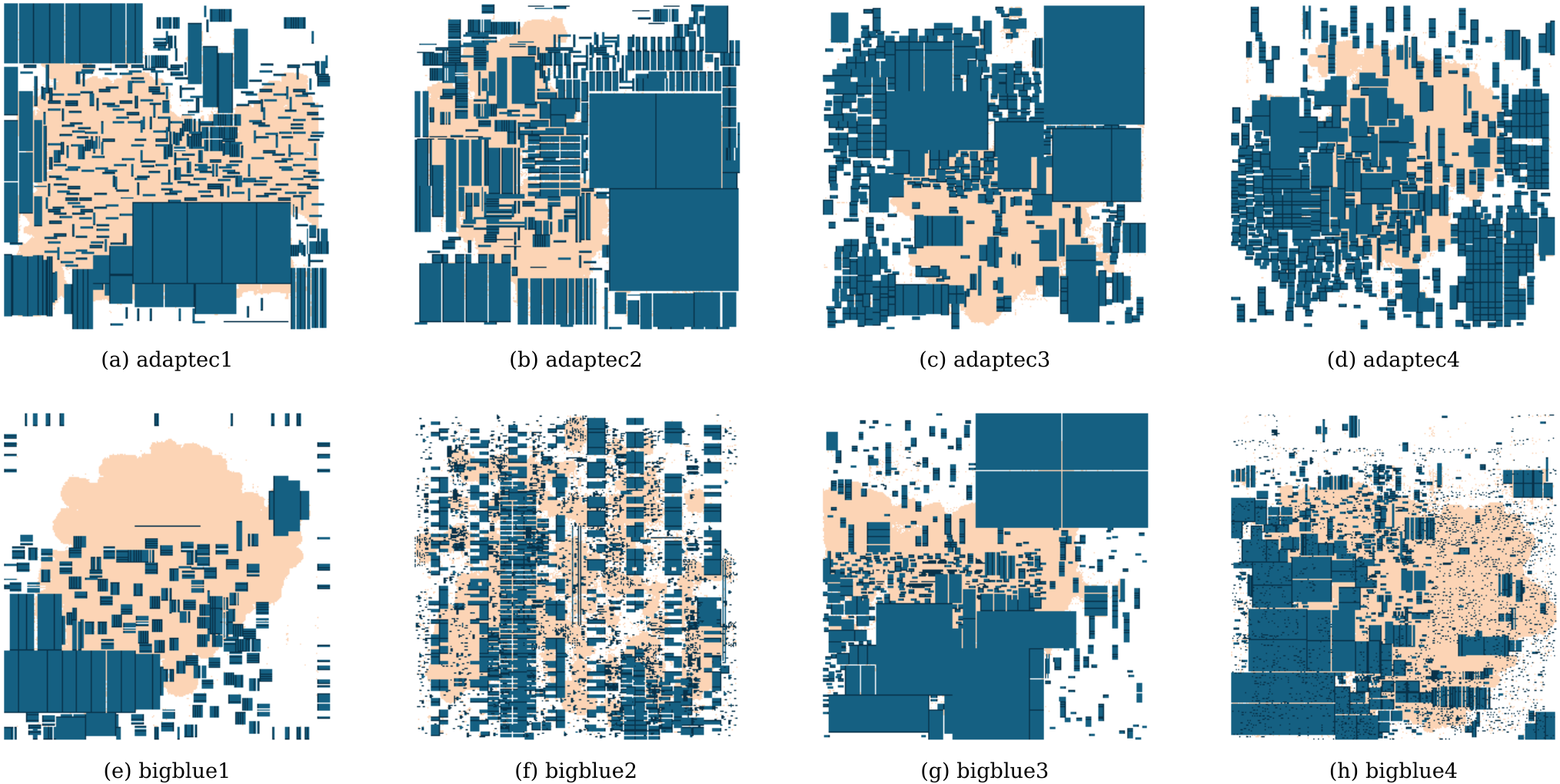}
\caption{
Visualization of mixed-size placement results for the ISPD2005 benchmarks in our evaluations.
Macros are shown in \textcolor[HTML]{156082}{blue}, and standard cells are shown in \textcolor[HTML]{FCD4B5}{orange}.}
\label{fig:appendix-mixed-size-placements}
\end{figure*}

\paragraph{Congestion evaluation.}
We further investigate the downstream congestion results using RUDY overflow~\citep{spindler2007rudy}. 
The detailed evaluation settings are reported above, and EfficientPlace and DiversePlace use the same routing-bin resolution for each benchmark. 
Table~\ref{tab:appendix-congestion} reports the normalized congestion comparison, where EfficientPlace is normalized to $1.00$ on each dataset and lower values are better. 
Although our primary optimization objective is HPWL, DiversePlace achieves lower normalized RUDY overflow across the evaluated circuits. 
These results indicate that the quality-guided diverse archive can have a positive effect on downstream routability, suggesting that the additional layout diversity is useful beyond improving archive variation alone.

\begin{table*}[!htbp]
\centering
\scriptsize
\setlength{\tabcolsep}{3.0pt}
\caption{
Normalized RUDY congestion comparison.
Lower values are better.
The MaskPlace row reports the normalized reference values from~\cite{geng2024efficientplace} for reference only.
}
\label{tab:appendix-congestion}
\resizebox{\textwidth}{!}{%
\begin{tabular}{@{}C{3.0cm}*{8}{C{1.25cm}}@{}}
\toprule
Method & adaptec1 & adaptec2 & adaptec3 & adaptec4 & bigblue1 & bigblue2 & bigblue3 & bigblue4 \\
\midrule
MaskPlace reference & 1.17 & 2.09 & 1.18 & 1.20 & 4.43 & -- & 1.70 & -- \\
Top-5 EfficientPlace & 1.00 & 1.00 & 1.00 & 1.00 & 1.00 & 1.00 & 1.00 & 1.00 \\
Top-5 DiversePlace & \textbf{0.57} & \textbf{0.94} & \textbf{0.75} & \textbf{0.90} & \textbf{0.70} & \textbf{0.73} & \textbf{0.74} & \textbf{0.80} \\
\bottomrule
\end{tabular}%
}
\end{table*}

\FloatBarrier
\section{Sensitivity Analysis of Key Parameters}
\label{app:sensitivity-analysis}

The gated quality threshold and QD-map resolution are the two key hyperparameters that directly control our quality-guided diversity mechanism.
The gated quality threshold determines which candidate placements are considered competitive enough to enter the diversity-seeking archive, while the QD-map resolution controls how finely the archive separates different displacement modes.
We conduct a focused sensitivity study on \textit{adaptec1}, using HPWL as the placement-quality metric and $D_{20}^{\mathrm{full}}$ and $D_{20}^{\mathrm{PPO}}$ as the archive-diversity metrics, same as in Section~\ref{sec:experiments}.
All sensitivity rows use cached top-20 placement archives; the ``DiversePlace default'' row uses our main-experiment setting, with gated quality epsilon $0.05$ and $8$ QD-map bins, reporting rounded values from Tables~\ref{tab:hpwl-overall} and~\ref{tab:diversity}.

\noindent
\parbox[t]{0.49\linewidth}{%
\centering
\captionof{table}{Gated quality $\epsilon$ sensitivity on \textit{adaptec1}.}
\label{tab:appendix-epsilon-sensitivity}
{\scriptsize
\setlength{\tabcolsep}{2.0pt}
\scalebox{1.2}{%
\begin{tabular}{@{}cccc@{}}
\toprule
$\epsilon$ & HPWL & $D_{20}^{\mathrm{full}}$ & $D_{20}^{\mathrm{PPO}}$ \\
\midrule
0.03 & 5.95$\pm$0.06 & \textbf{3.99} & 1.70 \\
\textbf{0.05 default} & \textbf{5.91}$\pm$0.05 & 3.86 & 1.81 \\
0.07 & 5.93$\pm$0.07 & 3.98 & \textbf{1.85} \\
0.10 & 5.97$\pm$0.06 & 3.84 & 1.67 \\
\bottomrule
\end{tabular}%
}
}
}
\hfill
\parbox[t]{0.49\linewidth}{%
\centering
\captionof{table}{QD-map bin sensitivity on \textit{adaptec1}.}
\label{tab:appendix-qdbin-sensitivity}
{\scriptsize
\setlength{\tabcolsep}{2.0pt}
\scalebox{1.2}{%
\begin{tabular}{@{}cccc@{}}
\toprule
QD bins & HPWL & $D_{20}^{\mathrm{full}}$ & $D_{20}^{\mathrm{PPO}}$ \\
\midrule
6 & \textbf{5.88}$\pm$0.02 & 3.60 & 1.52 \\
\textbf{8 default} & 5.91$\pm$0.05 & 3.86 & \textbf{1.81} \\
10 & 5.94$\pm$0.05 & \textbf{3.93} & 1.70 \\
12 & 5.98$\pm$0.09 & 3.86 & 1.70 \\
\bottomrule
\end{tabular}%
}
}
}
\vspace{3pt}

Table~\ref{tab:appendix-epsilon-sensitivity} shows that relaxing the gate does not monotonically improve the archive.
A tighter $\epsilon=0.03$ gate yields the highest full-placement diversity, while $\epsilon=0.07$ gives the highest PPO-region diversity.
The default $\epsilon=0.05$ achieves the best mean HPWL with competitive diversity.
Further relaxation to $\epsilon=0.10$ degrades HPWL, suggesting that admitting lower-quality candidates to the diversity phase can weaken placement optimization.
We use $\epsilon=0.05$ across datasets as a conservative default that preserves the quality filter and encourages nontrivial diversity within the archive.

Table~\ref{tab:appendix-qdbin-sensitivity} shows the expected tradeoff between map granularity and optimization pressure.
Six bins give the best HPWL on \textit{adaptec1} but reduce both full-placement and PPO-region diversity relative to the default.
Ten or twelve bins encourage broader archive coverage, although stronger diversity pressure can degrade HPWL.
The default eight bins balance strong HPWL with the highest PPO-region diversity among the tested settings.
We therefore fix eight bins across datasets in all main experiments rather than select a circuit-specific value.
The six-bin result suggests additional headroom from per-instance tuning, but we report the uniform default to maintain a fair and reusable protocol for evaluation across datasets.

\FloatBarrier
\section{Mixed-Size Visualizations for Diversity Comparison}
\label{app:mixed-size-diversity}

Figure~\ref{fig:appendix-adaptec4-mixed-size-top5-archive} extends the macro-only archive visualization in Figure~\ref{fig:adaptec4-top5-archive} to the corresponding mixed-size placements.
We use exactly the same \textit{adaptec4} macro candidates as Figure~\ref{fig:adaptec4-top5-archive}; after fixing those macro locations, DREAMPlace places the standard cells and refines the complete mixed-size layout.
Figure~\ref{fig:appendix-adaptec3-mixed-size-top5-archive} applies the same visualization protocol to \textit{adaptec3}, providing an additional example of how archive-level macro diversity carries over after mixed-size placement.
Together, these views show whether the macro-level archive diversity remains visible after standard-cell insertion.

\begin{figure}[!htbp]
\centering
\includegraphics[width=\textwidth]{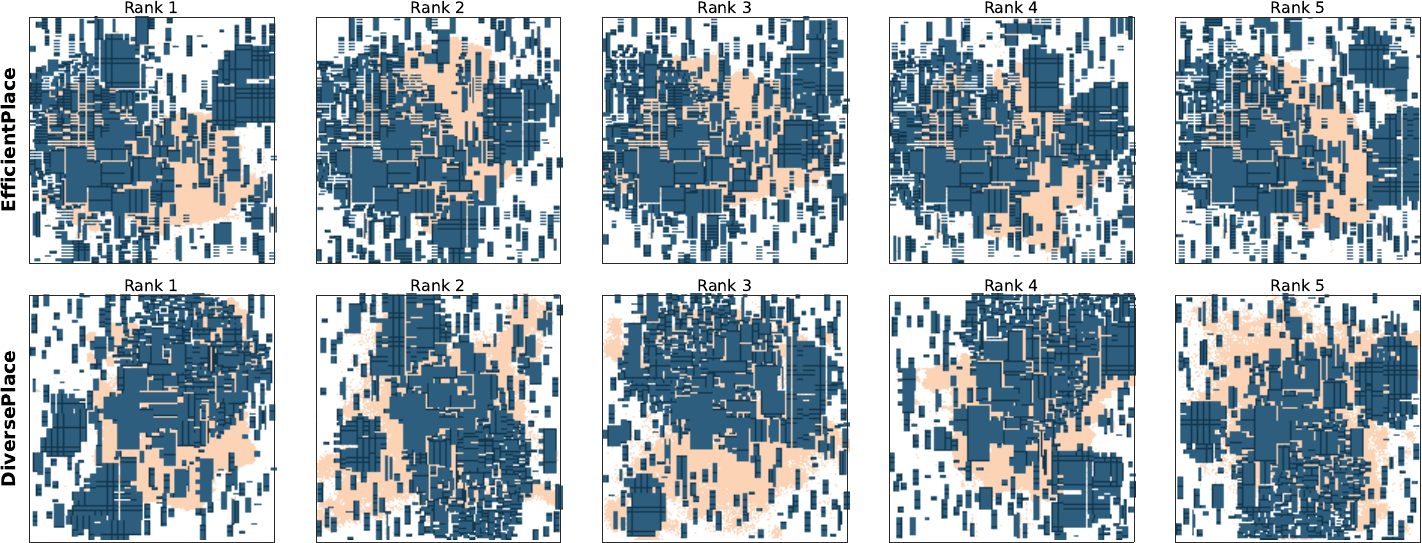}
\caption{
\textbf{Mixed-size version of Figure~\ref{fig:adaptec4-top5-archive} on \textit{adaptec4}.}
The macro candidates are exactly the top-5 archive placements shown in Figure~\ref{fig:adaptec4-top5-archive}.
Macros are shown in \textcolor[HTML]{2D5E7E}{blue} and standard cells in \textcolor[HTML]{FCD4B5}{orange}.}
\label{fig:appendix-adaptec4-mixed-size-top5-archive}
\end{figure}

\begin{figure}[!htbp]
\centering
\includegraphics[width=\textwidth]{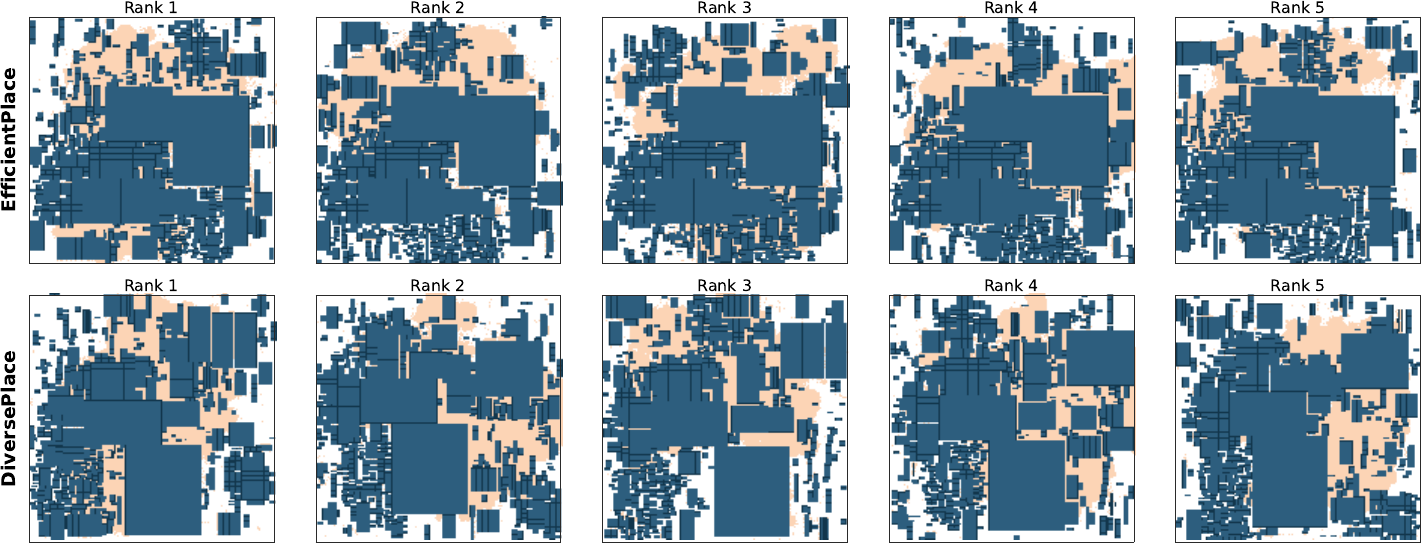}
\caption{
\textbf{Mixed-size top-5 archive placements on \textit{adaptec3}.}
The top and bottom rows show EfficientPlace and DiversePlace candidates, respectively.
Macros are shown in \textcolor[HTML]{2D5E7E}{blue}, and standard cells are shown in \textcolor[HTML]{FCD4B5}{orange}.}
\label{fig:appendix-adaptec3-mixed-size-top5-archive}
\end{figure}

\FloatBarrier
\section{Full Placement-and-Routing Evaluation}
\label{app:full-pr-evaluation}

\subsection{HPWL-Only Objective}
\label{app:ariane133-full-flow}

DiversePlace increases the empirical yield of high-quality downstream candidates by exploring distinct placement modes within a competitive HPWL region.
This directly addresses the proxy-driven nature of macro placement: preserving alternative layouts gives downstream implementation more opportunities to find favorable timing--wirelength solutions.

\paragraph{Evaluation protocol.}
We evaluate DiversePlace and EfficientPlace on \textit{ariane133} using the same official ORFS/OpenROAD flow.
Both methods use training seeds 0, 1, and 2, and every seed contributes equally to the reported results.
Each run supplies 20 HPWL-ranked candidates, giving both methods the same downstream evaluation budget.

\paragraph{High-quality candidate yield.}
DiversePlace completes the full flow for 98.33\% of its candidates, compared with 86.67\% for EfficientPlace.
More importantly, its jointly high-quality candidate yield increases from 18.33\% to 35.00\%.
A candidate meets this criterion when its Fmax is at least the median of EfficientPlace's successful candidates and its routed wirelength is at most the corresponding median.
These shared thresholds evaluate whether an archive provides candidates that are simultaneously strong in timing and wirelength.

\begin{table}[H]
\centering
\small
\setlength{\tabcolsep}{4pt}
\renewcommand{\arraystretch}{1.15}
\caption{Archive utility on \textit{ariane133} across seeds 0, 1, and 2, with 20 candidates per seed.
Joint-quality yield includes all evaluated candidates in the denominator.
Pareto counts are computed from the pooled successful candidates of both methods.}
\label{tab:ariane133-archive-utility}
\begin{tabular*}{\linewidth}{@{\extracolsep{\fill}}C{0.60\linewidth}C{0.17\linewidth}C{0.17\linewidth}@{}}
\toprule
Archive statistic & EfficientPlace & DiversePlace \\
\midrule
Full-flow completion & 52/60 (86.67\%) & \textbf{59/60 (98.33\%)} \\
Jointly high-quality candidates & 11/60 (18.33\%) & \textbf{21/60 (35.00\%)} \\
\mbox{Candidates on the common Fmax--wirelength Pareto frontier} & 3 & \textbf{12} \\
\bottomrule
\end{tabular*}
\end{table}

The nearly doubled joint-quality yield in Table~\ref{tab:ariane133-archive-utility} shows that DiversePlace supplies more useful candidates within the same evaluation budget.
Its larger contribution to the Pareto frontier further expands the available timing--wirelength choices.
Together, these results support the central purpose of the diverse archive: increasing the opportunity to discover a strong downstream implementation through broader coverage of competitive placement modes.

\paragraph{Downstream selection across runs.}
This higher candidate yield is accompanied by better selected implementations when results are averaged over all three seeds.
We compare selecting the lowest-HPWL candidate with selecting the highest-Fmax successful implementation from the Top-20 archive.
For each seed, all reported metrics come from the same selected placement.

\begin{table}[H]
\centering
\small
\setlength{\tabcolsep}{3pt}
\renewcommand{\arraystretch}{1.15}
\caption{Post-route results on \textit{ariane133}, averaged equally over seeds 0, 1, and 2.
All metrics within a seed are measured on the same selected placement.}
\label{tab:ariane133-full-flow-selection}
\begin{tabular*}{\linewidth}{@{\extracolsep{\fill}}C{0.16\linewidth}C{0.18\linewidth}*{4}{C{0.14\linewidth}}@{}}
\toprule
Selection & Method & \shortstack{Fmax $\uparrow$\\(MHz)} & \shortstack{Setup WNS $\uparrow$\\(ns)} & \shortstack{Setup TNS $\uparrow$\\(ns)} & \shortstack{Routed WL $\downarrow$\\($\times10^6$)} \\
\midrule
HPWL rank-1 & EfficientPlace & 182.25 & $-1.49$ & $-92.67$ & 6.12 \\
HPWL rank-1 & DiversePlace & \textbf{182.65} & $\mathbf{-1.48}$ & $\mathbf{-91.58}$ & 6.12 \\
\midrule
Top-20 Fmax & EfficientPlace & 183.73 & $-1.44$ & $-89.75$ & 6.13 \\
Top-20 Fmax & DiversePlace & \textbf{184.19} & $\mathbf{-1.43}$ & $\mathbf{-88.74}$ & \textbf{6.11} \\
\bottomrule
\end{tabular*}
\end{table}

Selecting from the DiversePlace archive improves average Fmax, setup slack, and routed wirelength relative to HPWL rank-1 selection.
DiversePlace also achieves better mean timing than EfficientPlace under both selection rules and shorter routed wirelength with Top-20 Fmax selection (Table~\ref{tab:ariane133-full-flow-selection}).
The combination of higher candidate yield and stronger cross-seed selection results demonstrates the practical benefit of preserving multiple competitive layouts for downstream evaluation.

\subsection{Composite Quality Objective}
\label{app:composite-quality-full-flow}

We further evaluate whether DiversePlace can provide useful downstream alternatives when placement quality incorporates regularity and logical grouping alongside HPWL.
Motivated by regularity-aware refinement in MaskRegulate~\citep{xue2024macroregulator} and expert-guided macro grouping in EXPlace~\citep{gao2026expertise}, we use the composite objective
\begin{equation}
Q(P)=0.8\,\widetilde H(P)+0.1\,\widetilde R(P)+0.1\,\widetilde G(P),
\label{eq:appendix-composite-quality}
\end{equation}
where the three terms measure normalized HPWL, boundary regularity, and hierarchy-based group compactness.
Each cost is normalized by its value on the benchmark's fixed legalized reference placement.
Both methods use the same composite objective and feasible-action constraints, and candidates are ranked by $Q$.

\paragraph{Evaluation across seeds and archive sizes.}
We use the same training seeds, 0, 1, and 2, for both methods and include all three in the reported mean and sample standard deviation.
For each seed, we evaluate the composite-ranked Top-5, Top-10, and Top-20 archives through the same ORFS/OpenROAD flow described above.
At each archive size, we select the successfully routed candidate with the highest Fmax and report all metrics from that placement.

\begin{table}[H]
\centering
\small
\setlength{\tabcolsep}{3pt}
\renewcommand{\arraystretch}{1.15}
\caption{Post-route results from composite-ranked archives on \textit{ariane133}.
Entries are mean $\pm$ sample standard deviation over seeds 0, 1, and 2.
All metrics within a seed and archive size come from the same Fmax-selected placement.
Relative improvements are calculated before display rounding.}
\label{tab:composite-quality-full-flow}
\begin{tabular*}{\linewidth}{@{\extracolsep{\fill}}C{0.12\linewidth}C{0.17\linewidth}*{4}{C{0.155\linewidth}}@{}}
\toprule
Archive & Method & \shortstack{Fmax $\uparrow$\\(MHz)} & \shortstack{Setup WNS $\uparrow$\\(ns)} & \shortstack{Setup TNS $\uparrow$\\(ns)} & \shortstack{Routed WL $\downarrow$\\($\times10^6$)} \\
\midrule
Top-5 & EfficientPlace & $181.74\pm5.34$ & $-1.51\pm0.16$ & $-94.25\pm9.88$ & $7.54\pm0.45$ \\
Top-5 & DiversePlace & $\mathbf{184.05\pm0.47}$ & $\mathbf{-1.43\pm0.01}$ & $\mathbf{-89.37\pm1.06}$ & $\mathbf{6.00\pm0.06}$ \\
\midrule
Top-10 & EfficientPlace & $182.16\pm5.83$ & $-1.49\pm0.18$ & $-93.22\pm11.06$ & $7.54\pm0.45$ \\
Top-10 & DiversePlace & $\mathbf{184.18\pm0.42}$ & $\mathbf{-1.43\pm0.01}$ & $\mathbf{-89.21\pm1.03}$ & $\mathbf{6.02\pm0.08}$ \\
\midrule
Top-20 & EfficientPlace & $182.20\pm5.87$ & $-1.49\pm0.18$ & $-93.37\pm10.88$ & $7.51\pm0.42$ \\
Top-20 & DiversePlace & $\mathbf{184.18\pm0.42}$ & $\mathbf{-1.43\pm0.01}$ & $\mathbf{-89.21\pm1.03}$ & $\mathbf{6.02\pm0.08}$ \\
\bottomrule
\end{tabular*}
\end{table}

\paragraph{Consistent downstream quality.}
DiversePlace achieves higher mean Fmax, better setup slack, and shorter routed wirelength at every evaluated archive size (Table~\ref{tab:composite-quality-full-flow}).
Across Top-5, Top-10, and Top-20, it improves mean Fmax by 1.09--1.27\% and reduces routed wirelength by 19.83--20.40\%.
These gains are accompanied by lower variation across runs: its Fmax standard deviation is 0.42--0.47~MHz, compared with 5.34--5.87~MHz for EfficientPlace.

The favorable mean results across archive sizes, together with the smaller across-seed variation, show that DiversePlace provides consistent access to strong downstream implementations under the composite objective.
This extends the archive-based formulation to a richer quality definition: quality-guided exploration across distinct placement modes continues to produce useful timing and wirelength choices across repeated runs.

\section{Broader Impacts}
\label{app:broader-impact}

This work studies RL-based macro placement for physical design.
Its primary positive impact is to improve the efficiency and quality of chip-design automation.
By producing multiple HPWL-competitive macro placements rather than a single proxy-optimal solution, DiversePlace can offer downstream physical-design tools a richer candidate set, potentially reducing manual design iterations and improving the chance of finding layouts with favorable mixed-size placement or congestion behavior.
More efficient macro placement may lower computing and engineering costs in early chip design, benefiting academic research and industrial electronic design automation workflows.

\clearpage
\endgroup


\end{document}